\documentclass{ieeeaccess}
\usepackage{cite}
\usepackage{amsmath,amssymb,amsfonts}

\usepackage{setspace}
\usepackage{caption}
\usepackage{subcaption}
\usepackage{algorithm}
\usepackage{algpseudocode}
\usepackage{url}

\usepackage{graphicx}
\usepackage{textcomp}
\def\BibTeX{{\rm B\kern-.05em{\sc i\kern-.025em b}\kern-.08em
    T\kern-.1667em\lower.7ex\hbox{E}\kern-.125emX}}

\begin{document}
\history{Date of publication xxxx 00, 0000, date of current version xxxx 00, 0000.}
\doi{10.0000/ACCESS.2000.DOI}

\title{The Effect of Points Dispersion on the $k$-nn Search in Random Projection Forests}
\author{\uppercase{Mashaan Alshammari}\authorrefmark{1},
\uppercase{John Stavrakakis\authorrefmark{2}},
\uppercase{Adel F. Ahmed\authorrefmark{3}},
\uppercase{Masahiro Takatsuka\authorrefmark{2}}.
}
\address[1]{Independent Researcher, Riyadh, Saudi Arabia (e-mail: mashaan.awad1930@alum.kfupm.edu.sa)}
\address[2]{School of Computer Science, University of Sydney, Sydney, 
NSW, Australia (e-mail: john.stavrakakis@sydney.edu.au; masa.takatsuka@sydney.edu.au)}
\address[3]{Information and Computer Science Department, King Fahd University of Petroleum and Minerals, Dhahran, Saudi Arabia (e-mail: adelahmed@kfupm.edu.sa)}

\markboth
{Alshammari \headeretal: Effect of Points Dispersion on the $k$-nn Search in Random Projection Forests}
{Alshammari \headeretal: Effect of Points Dispersion on the $k$-nn Search in Random Projection Forests}

\corresp{Corresponding author: Mashaan Alshammari (e-mail: mashaan.awad1930@alum.kfupm.edu.sa).}

\begin{abstract}
Partitioning trees are efficient data structures for $k$-nearest neighbor search. Machine learning libraries commonly use a special type of partitioning trees called $k$d-trees to perform $k$-nn search. Unfortunately, $k$d-trees can be ineffective in high dimensions because they need more tree levels to decrease the vector quantization (VQ) error. Random projection trees rpTrees solve this scalability problem by using random directions to split the data. A collection of rpTrees is called rpForest. $k$-nn search in an rpForest is influenced by two factors: 1) the dispersion of points along the random direction and 2) the number of rpTrees in the rpForest. In this study, we investigate how these two factors affect the $k$-nn search with varying $k$ values and different datasets. We found that with larger number of trees, the dispersion of points has a very limited effect on the $k$-nn search. One should use the original rpTree algorithm by picking a random direction regardless of the dispersion of points.
\end{abstract}


\titlepgskip=-15pt

\maketitle

\section{Introduction}\label{Introduction}
$k$-nearest neighbor ($k$-nn) search is one of the fundamental tasks in machine learning. Given a set of $n$ points in $d$ dimensions $X = \{x_1, x_2, \cdots, x_n \}$, the $k$-nn search problem is defined as the problem of finding the subset of $X$ containing $\{x_1, x_2, \cdots, x_k \}$ nearest neighbors to a test point $x_i$ \cite{hart2000pattern}. Machine learning algorithms make decisions based on samples matching, which is usually done using $k$-nearest neighbor search. It has been used in several machine learning tasks such as time series classification \cite{Gweon2021Nearest,Kowsar2022Shape} and spectral clustering \cite{Alshammari2021Refining,Yuan2020Spectral,Kim2021KNN}. 

In their extensive review of $k$-nn search methods, Muja and Lowe \cite{Muja2014Scalable} identified three categories of $k$-nn search methods: partitioning trees, hashing techniques and neighboring graph techniques. Performing $k$-nearest neighbor search on partitioning trees has been the go-to option for most of machine learning libraries (e.g., scikit-learn library in python \cite{scikit-learn, sklearn_api} and knnsearch function in Matlab \cite{MATLABknnsearch}). These libraries implement a special data structure named k-dimensional tree (or $k$d-tree for short) \cite{Bentley1975Multidimensional,Friedman1977Algorithm,Ram2013Which}. When branching the $k$d-tree, the median point of one dimension divides the points into two parts. At the next branch, the next dimension is used to divide the remaining points into two branches. This process continues to partition median split of alternating dimension until the leaves of the tree contain only a few data points. This restricts the splits in $k$d-tree to the existing dimensions, which makes them vulnerable to the (curse of dimensionality). In other words, a $k$d-tree increasingly needs more tree levels to decrease the vector quantization (VQ) error and produce good partitioning \cite{Dasgupta2008Random,Freund2008Learning,Dasgupta2015Randomized}. The introduction of random projection trees (rpTrees) generalizes the concept of $k$d-tree. Instead of cutting along an existing dimension, the points are projected onto a random direction and split at a random split point. This concept eases the need for more dimensions as the number of points grows. Random projection forests (rpForests) were introduced to overcome the errors made by single rpTrees, and it was suggested that the aggregated results are more reliable \cite{Yan2018Nearest,Yan2021Nearest}.

There are two factors that affect the performance of $k$-nearest neighbor search in rpForests. First, the dispersion of points along the projected dimension. Intuitively, the more we spread the points along this random axis, the less likely we are separating two neighbors. The second factor that affects $k$-nn search in rpForests is the number of trees. If two neighbors were separated in one tree, there is a higher chance that they will be placed together in other trees.

The purpose of the article is to investigate what is the effect of maximizing the points dispersion on $k$-nn search in rpForests. The $k$-nn search results were evaluated using two metrics proposed by (Yan et al., 2018) \cite{Yan2018Nearest} and (Yan et al., 2021) \cite{Yan2021Nearest}. Our experiments revealed that the gains from attempting to maximize the dispersion of points, are outweighed by the growth of computations as the number of trees increases. We also found out that as the number of trees increases, maximizing the dispersion did not improve the quality of $k$-nn search in rpForests. When the number of trees is large ($T>20$), we are better off using the original rpTree algorithm, that is picking a random direction regardless of the dispersion and use it to split the points.

This paper is organized as follows: in the next section we review the methods used for $k$-nn, and the advancements on rpTrees. In section \ref{ProposedApproach}, we present our approach to perform $k$-nn search in rpForest, how did we construct rpForests and how the $k$-nn search results were aggregated from different trees. In section \ref{Experiments}, experiments were explained and discussed.
\section{Related work}
\label{RelatedWork}

$k$-nearest neighbors can be found via exact or approximate search. The difference is that approximate search is faster, but unlike exact search, it is not guaranteed to return the true nearest neighbors. In some cases, exact search can be faster, for example using time series data \cite{rakthanmanon2012searching}. In d-dimensional case, prestructuring is a common way to perform approximate $k$-nn search and reduce the computational burden \cite{hart2000pattern}. Prestructured $k$-nn search techniques can be classified into two categories: 1) searching hash-based indices, or 2) searching hierarchical tree indices \cite{Ram2013Which}.

Locality-sensitive hashing (LSH) is one of the widely-used methods in hashing category \cite{Andoni2008Near}. The idea behind LSH is basic, instead of searching the data space, the search takes place in a hash table because it is faster and efficient. A query point $p$ will get a hash key and fall onto a hash bucket within the search space, the neighbors in that bucket are returned by $k$-nn search. A number of studies have successfully applied LSH to their applications. A kernelized LSH has been introduced to accommodate kernel functions and visual features \cite{Kulis2009Kernelized}, LSH was used to detect multi-source outliers \cite{Ma2021Outlier} and to build a k-means-like clustering method \cite{Mau2021LSH}. In LSH, the $k$-nn search results are largely dependent on the content of the hash table. The challenge is how to design a hash function that accurately encodes the data into a hash table.

\begin{figure*}
	\centering
	\includegraphics[width=\textwidth,height=20cm,keepaspectratio]{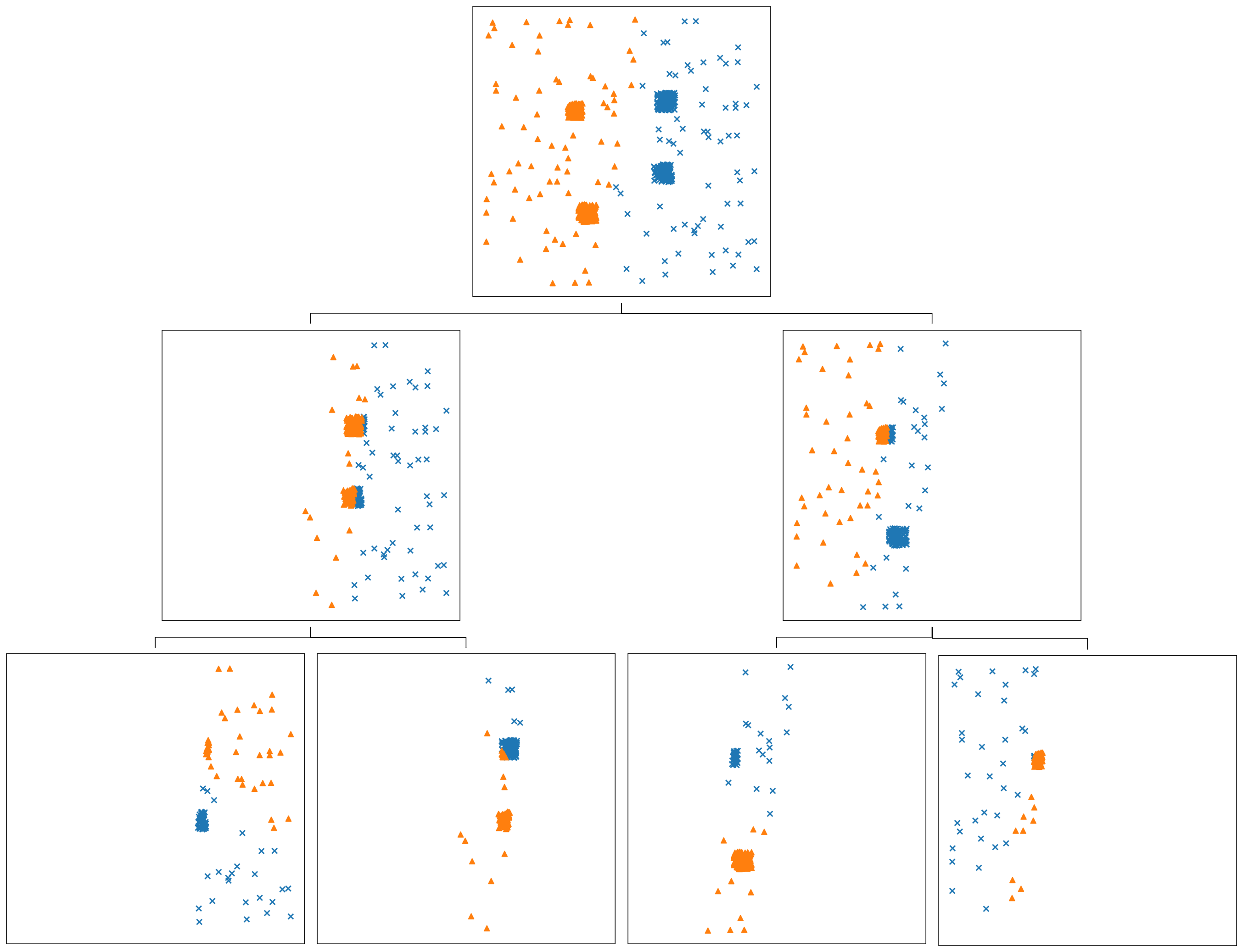}	
	\caption{An example of rpTree; points in blue are placed in the left child and points in orange are placed in the right child. (Best viewed in color)}
	\label{Fig:Fig-01}
\end{figure*}

Performing $k$-nearest neighbor search in partitioning trees overcomes the challenges posed by hash-based methods. Partitioning trees do not require a hash function and building the index is computationally efficient. $k$d-trees are widely used for $k$-nn search. However, the splits have to be done using the existing dimensions, which makes $k$d-trees vulnerable to the curse of dimensionality. rpTrees are a generalization of $k$d-trees, instead of splitting along an existing dimension, the split will occur along a random direction selected uniformly (as shown in Figure\ \ref{Fig:Fig-01}). Let $W$ be a node in an rpTree, all points in $W$ are projected onto a direction $\overrightarrow{r}$ chosen at random. Along the direction $\overrightarrow{r}$, a split point $c$ is selected randomly between $\lbrack \frac{1}{4}, \frac{3}{4} \rbrack$. Unlike the median split in $k$d-tree which occurs exactly at the median, $c$ is a perturbed split. If a projected point $x$ is less than $c$, it is placed in the left child $W_L$, otherwise, $x$ is placed in the right child $W_R$. Equation\ \ref{Eq-001} shows the placement rules in rpTrees.

\begin{equation}
	\begin{split}
		W_L={x \in W : x \cdot r < c} \\
		W_R={x \in W : x \cdot r \ge c}.
	\end{split}
	\label{Eq-001}
\end{equation}

The construction process of rpTree consists of two steps: 1) projecting the points in $W$ onto a random direction $\overrightarrow{r}$, and 2) picking a split point $c$ at random along $\overrightarrow{r}$. Ideally, these two steps should not separate two true neighbors into two different tree branches. To minimize this risk, (Yan et al., 2021) \cite{Yan2021Nearest} and (Yan et al., 2018) \cite{Yan2018Nearest} suggested that one should pick a number of random directions and project onto the one that yields the maximum dispersion of points. They also suggested using enough rpTrees in an rpForest to rectify errors produced by some rpTrees. However, the relationship between the dispersion of points ($\sigma_r$) and the number of trees ($T$) is not clear. In this work, we used four methods to construct rpForest. Each of which has its own strategy to handle the dispersion of points ($\sigma_r$).
\section{$k$-nearest neighbor search in random projection forests}
\label{ProposedApproach}

$k$-nearest neighbor search in random projection forests (rpForests) is influenced by two factors: the dispersion of points ($\sigma_r$) along a random direction and the number of rpTrees in rpForest ($T$). The relationship between these two factors was not covered in recent studies. Therefore, we designed our experiments to investigate how these two factors affect $k$-nn search in rpForests.

A single rpTree represents a block to build an rpForest. Performing $k$-nn search in rpForests involves: 1) building rpTrees $O(nlogn)$, 2) traversing the search sample $x$ from the root node of each tree to a leaf node $O(nlogn)$, and 3) aggregating the results across all trees and finding the nearest neighbors; where $n$ is the number of data points.

To investigate the relationship between the dispersion of points ($\sigma_r$) and the number of rpTrees in rpForest ($T$), we designed four methods to build rpTrees. The first method is the original rpTree algorithm developed by (Dasgupta and Freund, 2008) \cite{Dasgupta2008Random}. It picks a random direction regardless of the dispersion of points ($\sigma_r$). The second method of rpTree construction is the one developed by (Yan et al., 2021) \cite{Yan2021Nearest}. They suggested picking a number of random directions and use the one that yields the maximum dispersion of points. We developed the third method where we modified (Yan et al., 2021) \cite{Yan2021Nearest} method by introducing two further steps. In these steps, we rotate the direction with maximum dispersion to see if there is a larger dispersion of points. The last method is the optimal case, which is guaranteed to produce the maximum dispersion of points, that is applying principle component analysis (PCA) at each node of rpTree \cite{sproull1991refinements,Pearson1901On,abdi2010principal}.

\subsection{Building rpTrees}
\label{BuildingrpTrees}

Building an rpTree involves three main steps: 1) projecting $n$ data points in the current node $W$ onto a random direction $\overrightarrow{r}$, 2) picking a split point $c$ randomly between $1^{\text{st}}$ and $3^{\text{rd}}$ quantiles,
3) placing data points that are less than $c$ into the left child $W_L$ and all other points into the right child $W_R$, and 4) recursively perform steps 1 to 3, until the leaf nodes contain a number of points that are less than a predetermined integer. Algorithm \ref{Alg:Alg-01} shows the steps we used to construct the rpTrees.

\begin{algorithm}
	\caption{The recursive function used to construct $rpTrees$}\label{Alg:Alg-01}
	\begin{algorithmic}[1]
		\State Let $X$ be the set of points to be split in a tree $t$
		\Function{partition}{X}
		\If{$ \lvert X \rvert < n$}\label{algln2}
		\State \textbf{return} $makeNode(X)$
		\Else		
		\State \textcolor{blue}{// \textit{the used method selects $\overrightarrow{r}$}}
		\State Generate a random direction $\overrightarrow{r}$		
		\State Project points in $X$ onto $\overrightarrow{r}$
		\State Let $V$ be the projection of points in $X$ onto $\overrightarrow{r}$
		\State Pick $c \in V$ between $1^{\text{st}}$ and $3^{\text{rd}}$ quantiles.
		\State $X_L=\{x:V(x<c)\}$
		\State \textbf{return} $partition(X_L)$
		\State $X_R=\{x:V(x\ge c)\}$
		\State \textbf{return} $partition(X_R)$		
		\EndIf
		\EndFunction
	\end{algorithmic}
\end{algorithm}

\begin{figure}
	\centering
	\includegraphics[width=0.235\textwidth]{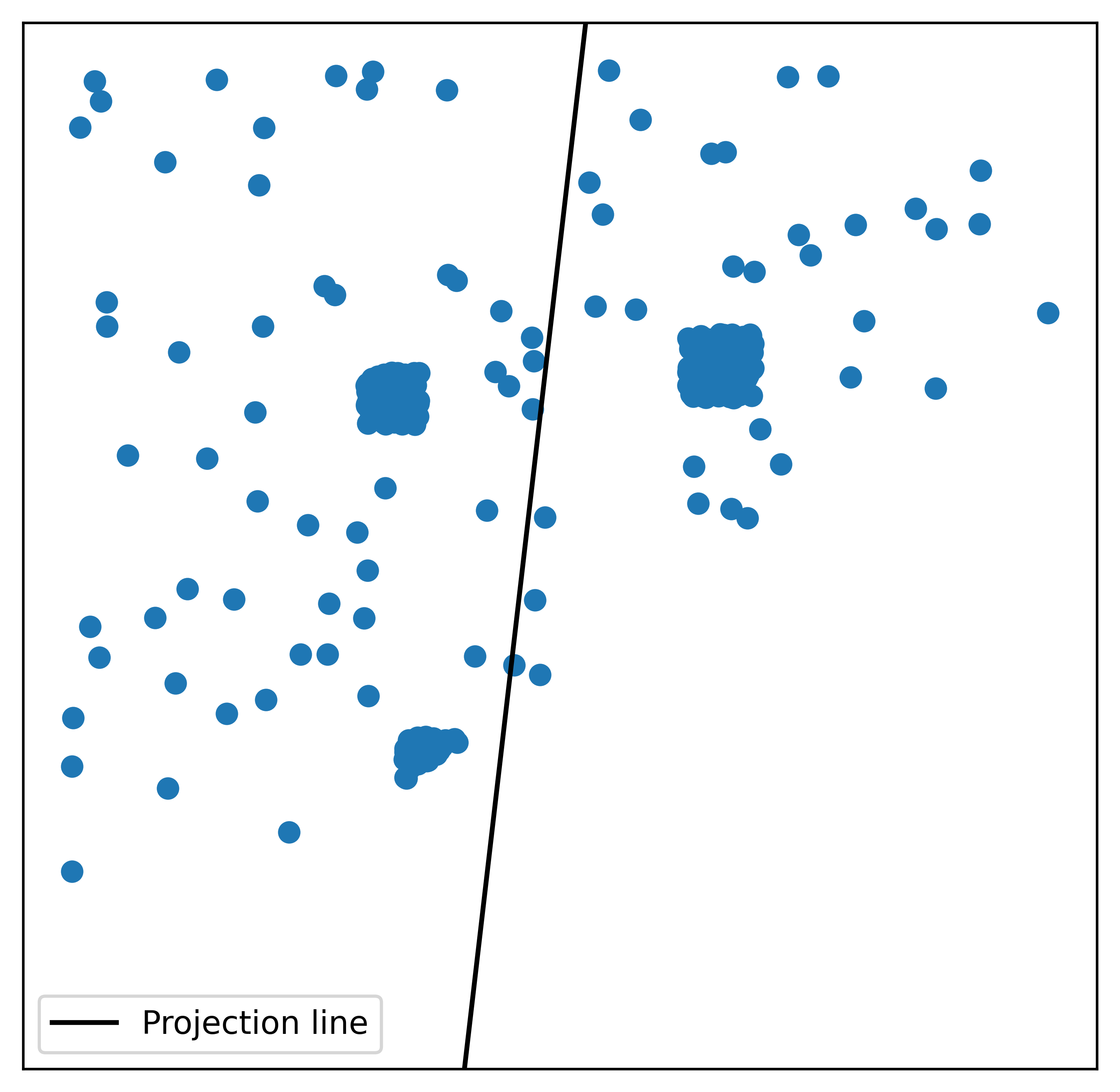}  
	\includegraphics[width=0.235\textwidth]{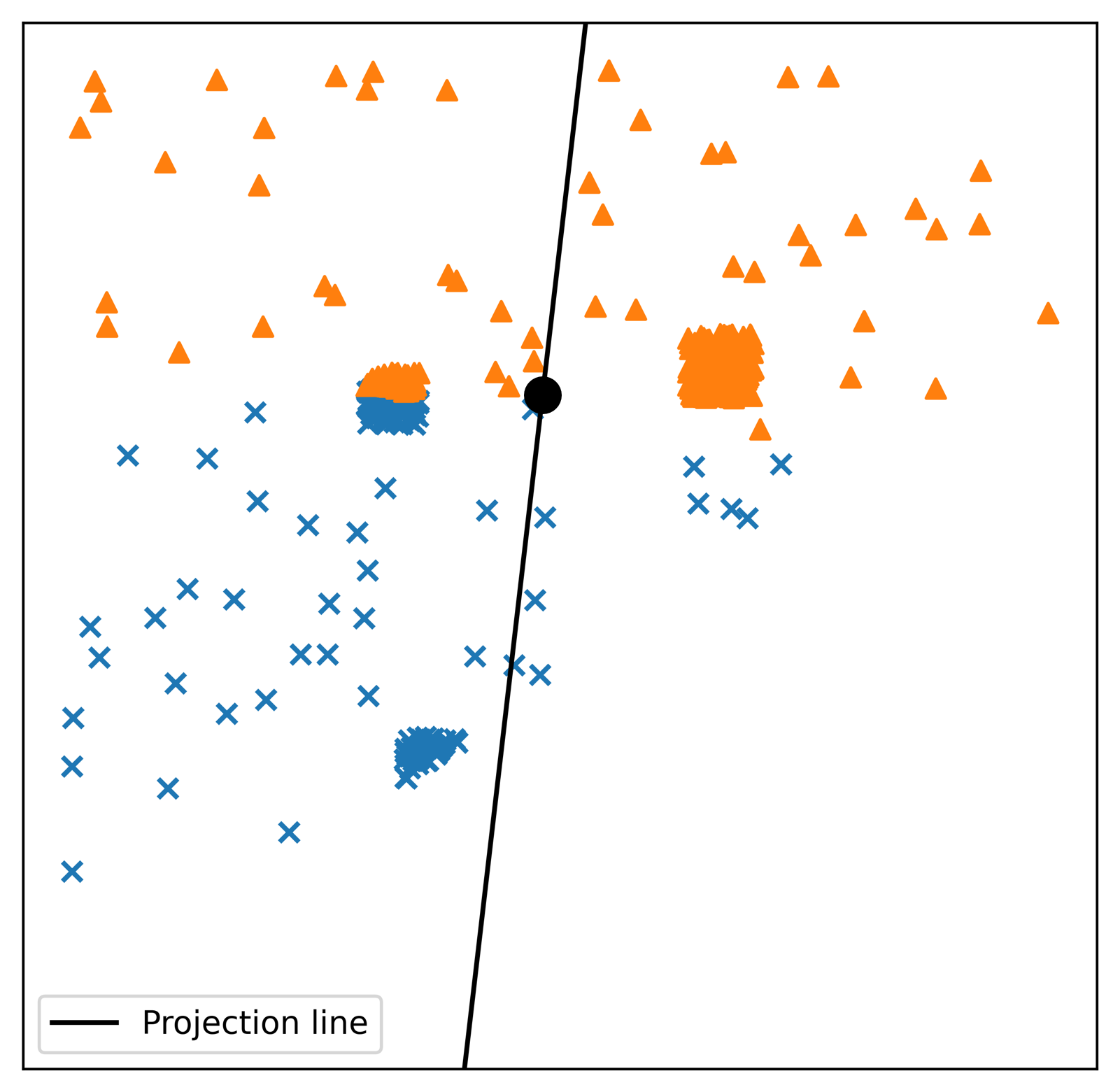}  
	\caption{A split made by \texttt{Method 1}; (left) picking a random direction; (right) splitting the points along that direction; the black point along the projection line is the split point. (Best viewed in color)}
	\label{Fig:Fig-02}
\end{figure}

The original algorithm of rpTree (which will be referred to as \texttt{Method 1} for the rest of this paper) suggests picking a random direction regardless of the dispersion of points. It is the fastest method in our experiments because it uses random projection only once and move to the second step of splitting the data points. Figure\ \ref{Fig:Fig-02} illustrates how this method performs projection and splitting of the points in a node $W$. Figure\ \ref{Fig:Fig-02} shows that \texttt{Method 1} projects completely at random, and the splits may not be perfect.

\begin{figure}
	\centering
	\includegraphics[width=0.235\textwidth]{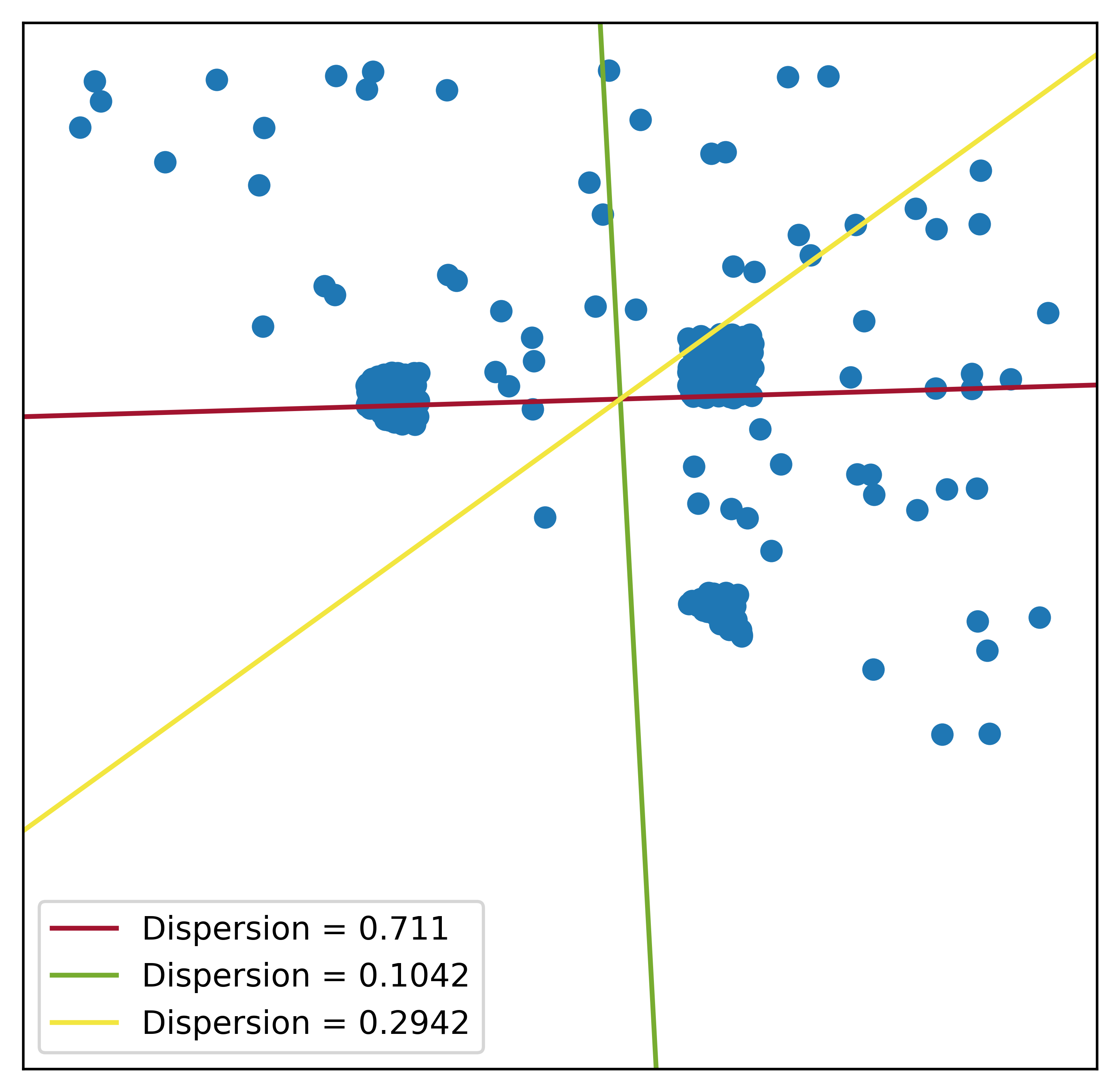}  
	\includegraphics[width=0.235\textwidth]{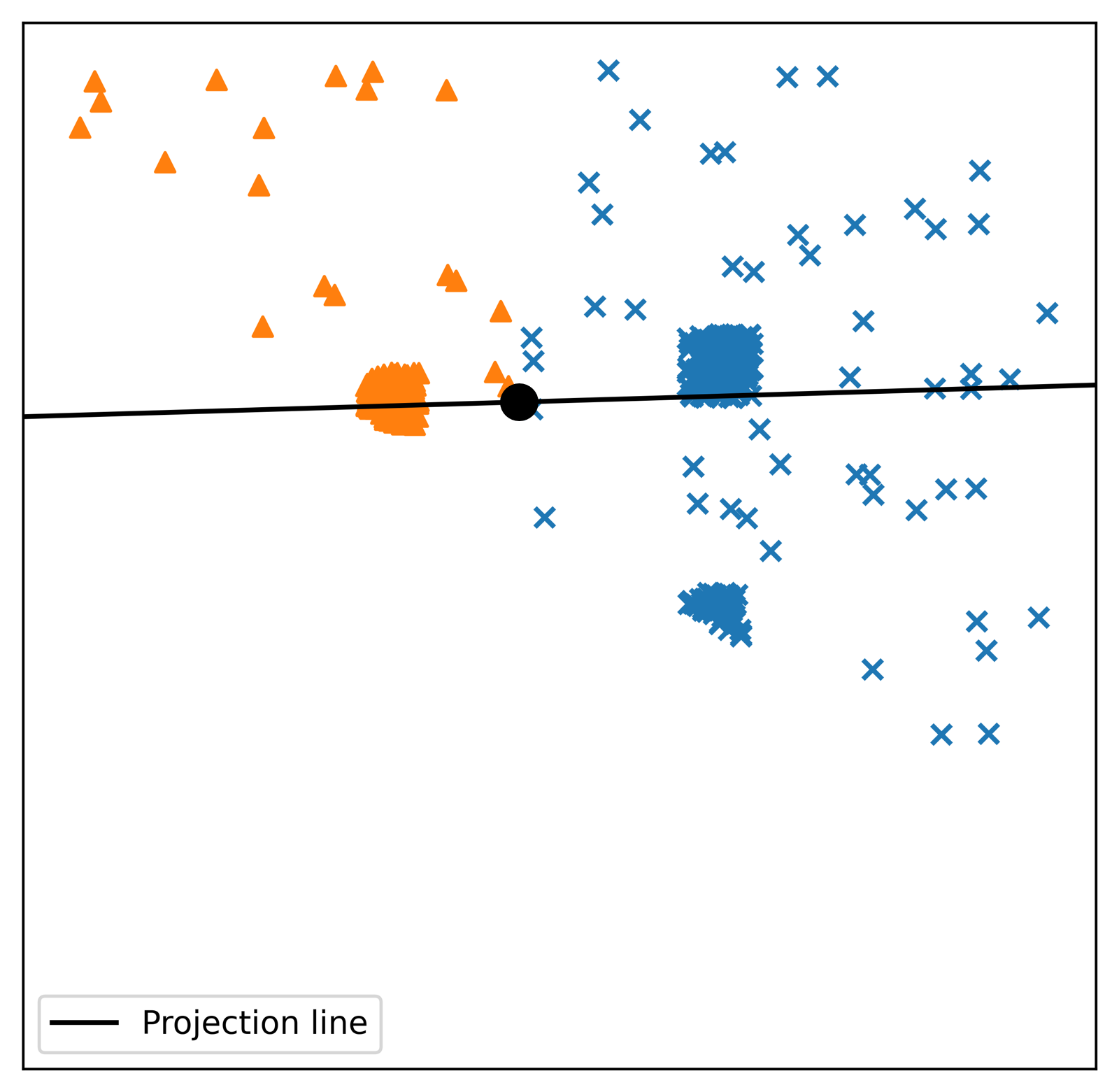}  
	\caption{\texttt{Method 2} chooses the projection dimension with the maximum dispersion; (left) measuring the dispersion of points on three random directions; (right) splitting the points along the direction of maximum dispersion; the black point along the projection line is the split point. (Best viewed in color)}
	\label{Fig:Fig-03}
\end{figure}

\texttt{Method 2} in our experiments is the method developed by (Yan et al., 2021) \cite{Yan2021Nearest}. They suggested picking a number of random directions (they used the symbol $nTry$) then use the one that yields the maximum dispersion of points. Apparently, it requires more time than \texttt{Method 1} because it performs random projection $nTry$ times not only once. An example of a node $W$ split by \texttt{Method 2} is shown in Figure\ \ref{Fig:Fig-03}.

\begin{figure*}
	\begin{subfigure}{0.24\textwidth}
	\centering
	\includegraphics[width=.98\linewidth]{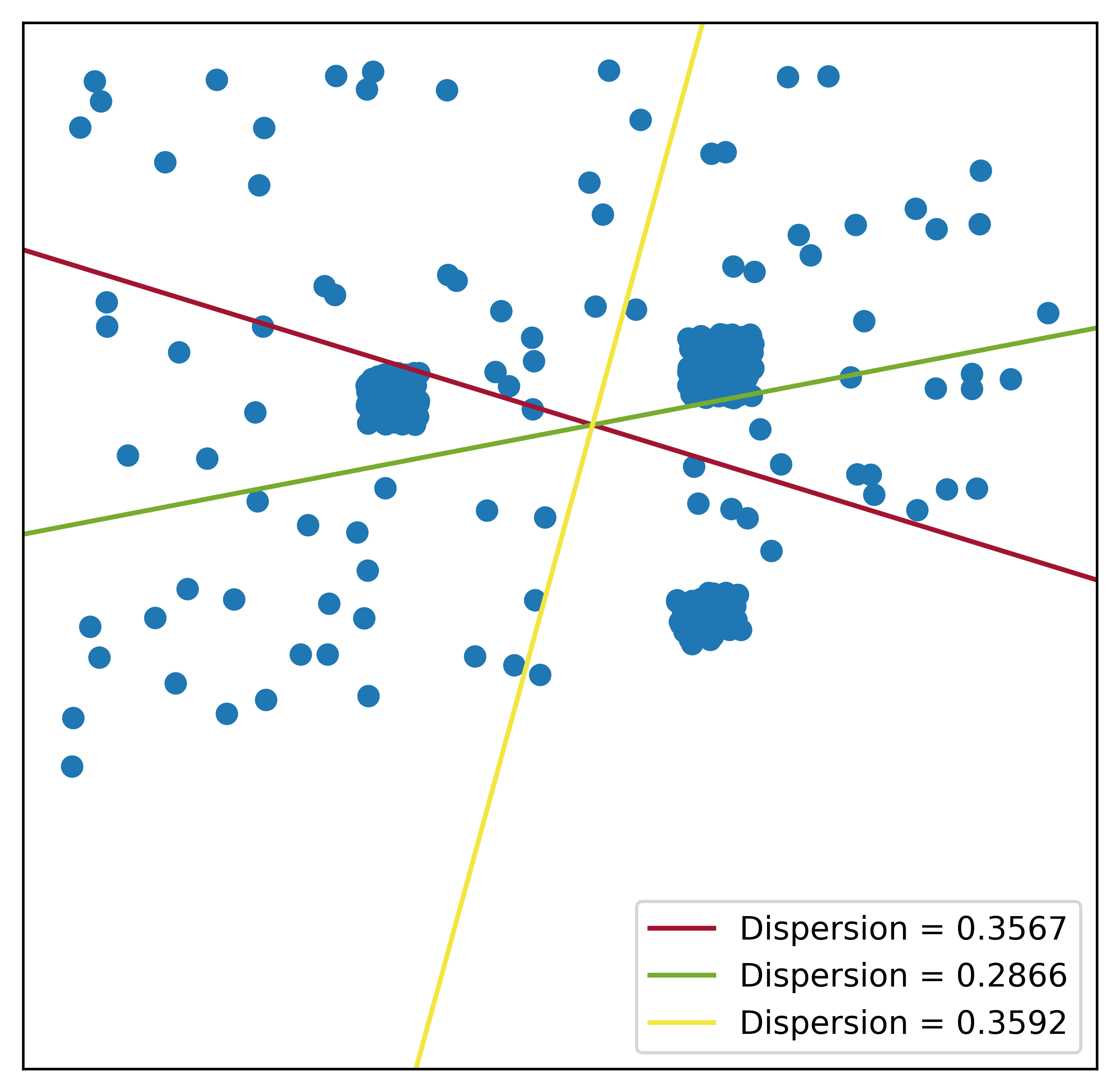}
	\caption{}
	\label{Fig:Fig-04-a}
	\end{subfigure}  
	\begin{subfigure}{0.24\textwidth}
	\centering
	\includegraphics[width=.98\linewidth]{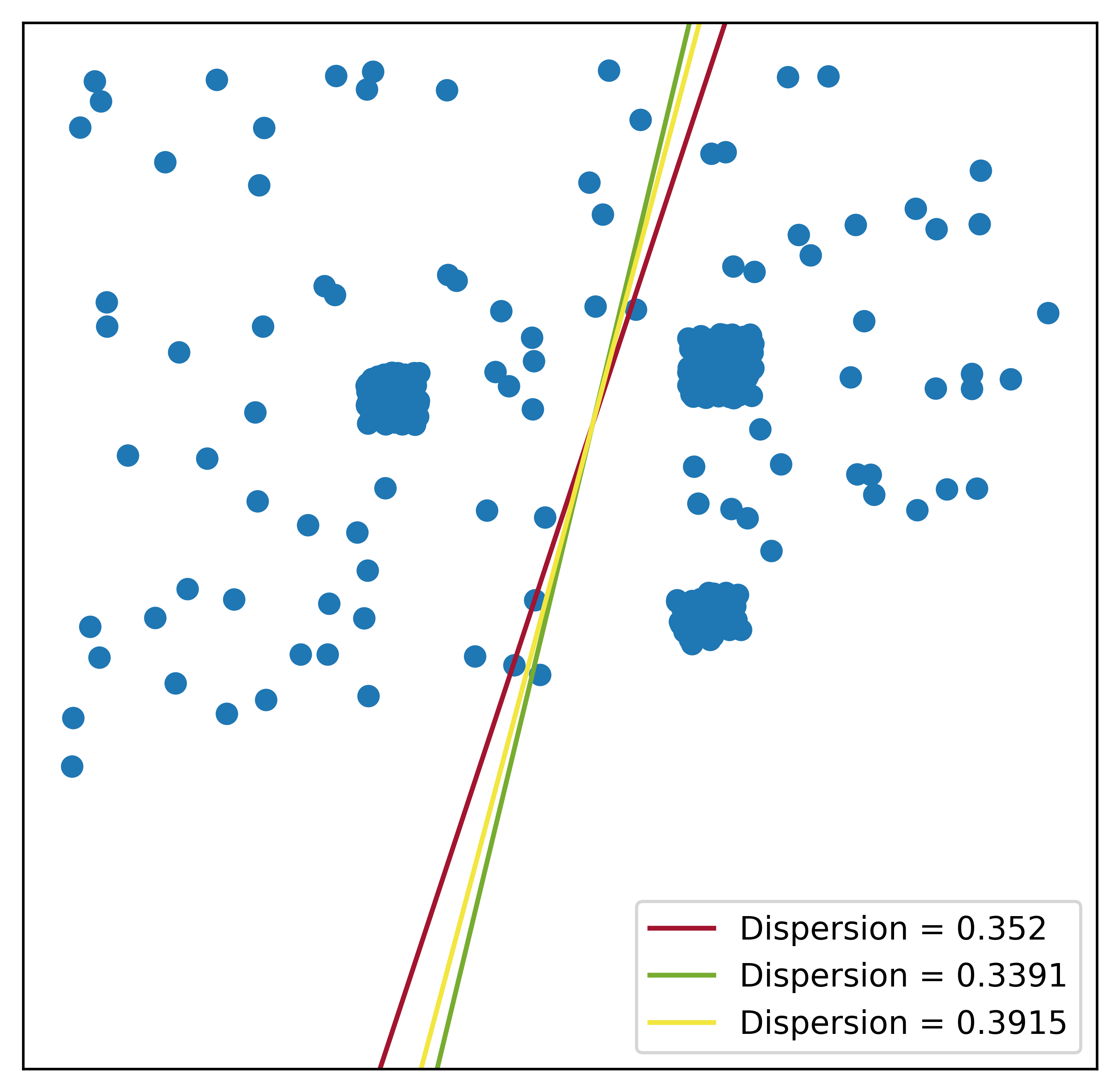}
	\caption{}
	\label{Fig:Fig-04-b}
	\end{subfigure}
	\begin{subfigure}{0.24\textwidth}
	\centering
	\includegraphics[width=.98\linewidth]{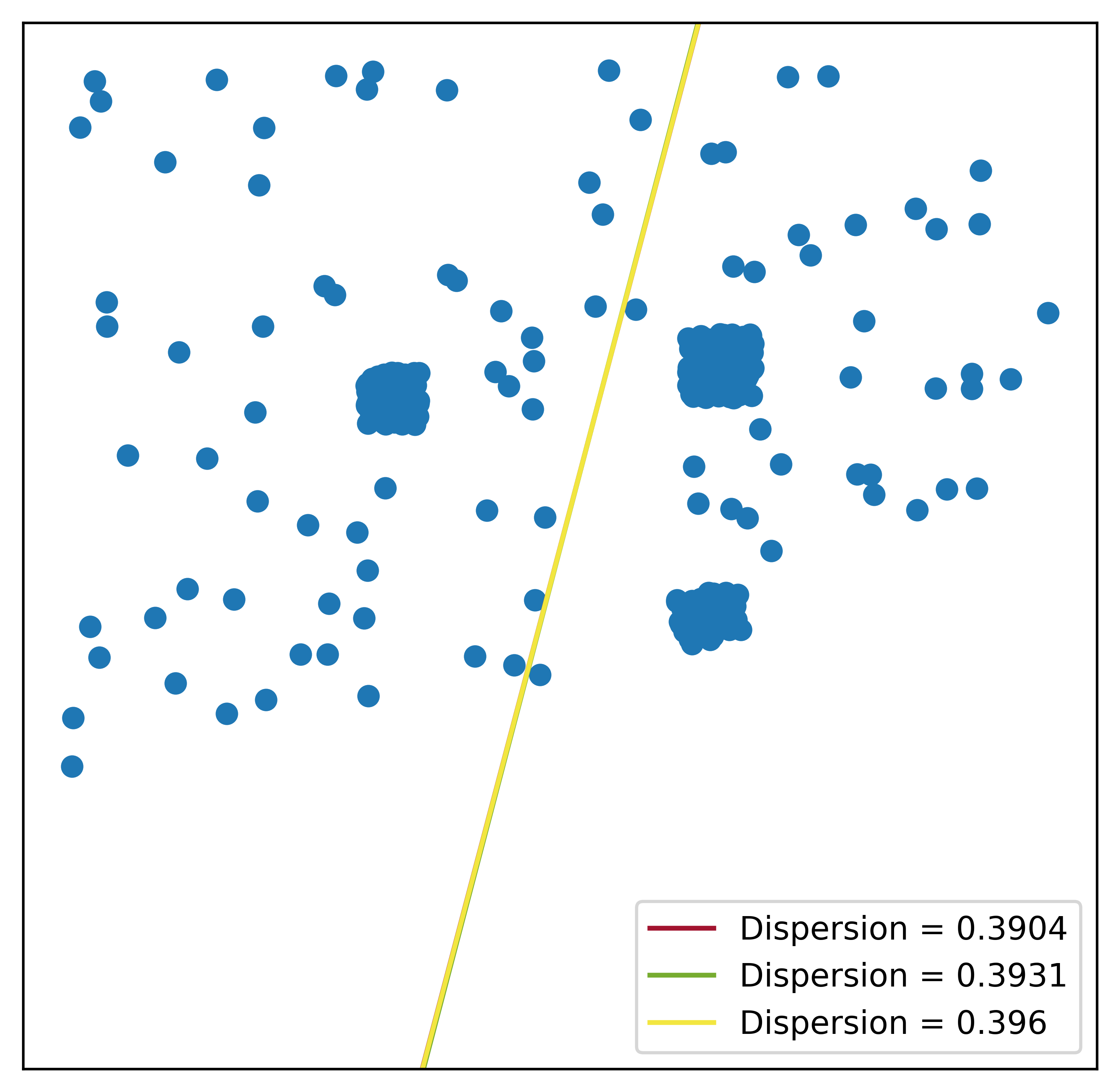}
	\caption{}
	\label{Fig:Fig-04-c}
	\end{subfigure}
	\begin{subfigure}{0.24\textwidth}
	\centering
	\includegraphics[width=.98\linewidth]{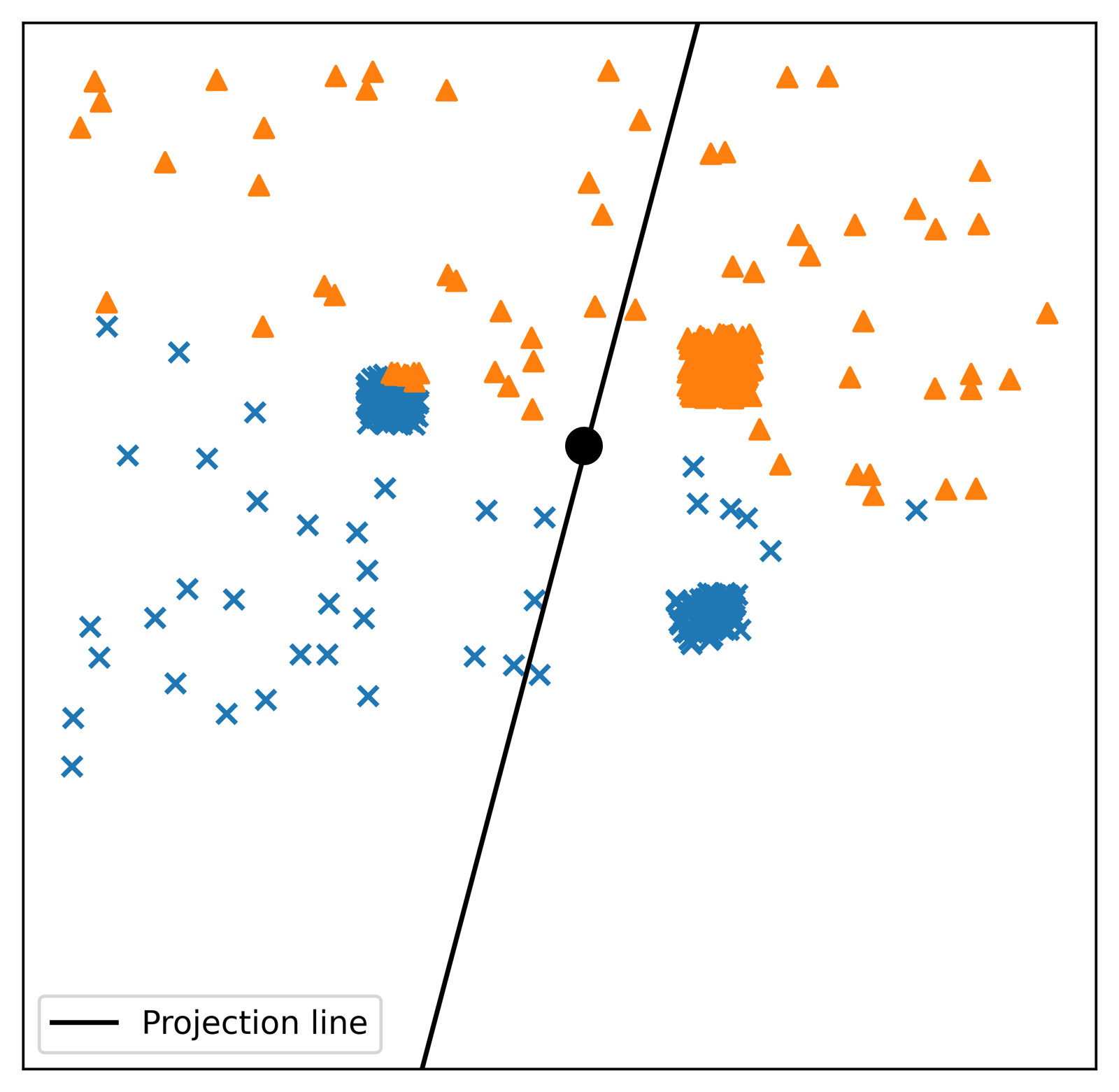}
	\caption{}
	\label{Fig:Fig-04-d}
	\end{subfigure}
	\caption{\texttt{Method 3} attempts to maximize the dispersion of points using three attempts; then it splits the points along the direction of maximum dispersion; the black point along the projection line is the split point. (Best viewed in color)}
	\label{Fig:Fig-04}
\end{figure*}

We developed \texttt{Method 3} in which we modified \texttt{Method 2}. Instead of just picking $nTry$ random directions, we extended this method by tuning the direction with the maximum dispersion. The tuning is done by adding Gaussian noise (with small $\sigma$) to the projection matrix, in order to maximize the dispersion. Unlike repeating the random projection, adding noise requires less computations. Figure\ \ref{Fig:Fig-04}. shows an example of this method. Initially, we picked three directions at random as in Figure\ \ref{Fig:Fig-04-a}. The yellow line has the maximum dispersion of points with $dispersion=0.3592$. The first step of direction tuning is adding noise with $\sigma=0.1$ to the projection matrix. This got us the yellow line with better dispersion of $0.3915$, shown in Figure\ \ref{Fig:Fig-04-b}. The second step of tuning is adding noise with $\sigma=0.01$ to the projection matrix. This gives us the winning direction shown in a yellow line with $dispersion=0.3960$, shown in Figure\ \ref{Fig:Fig-04-c}.

\begin{figure}
	\centering
	\includegraphics[width=0.235\textwidth]{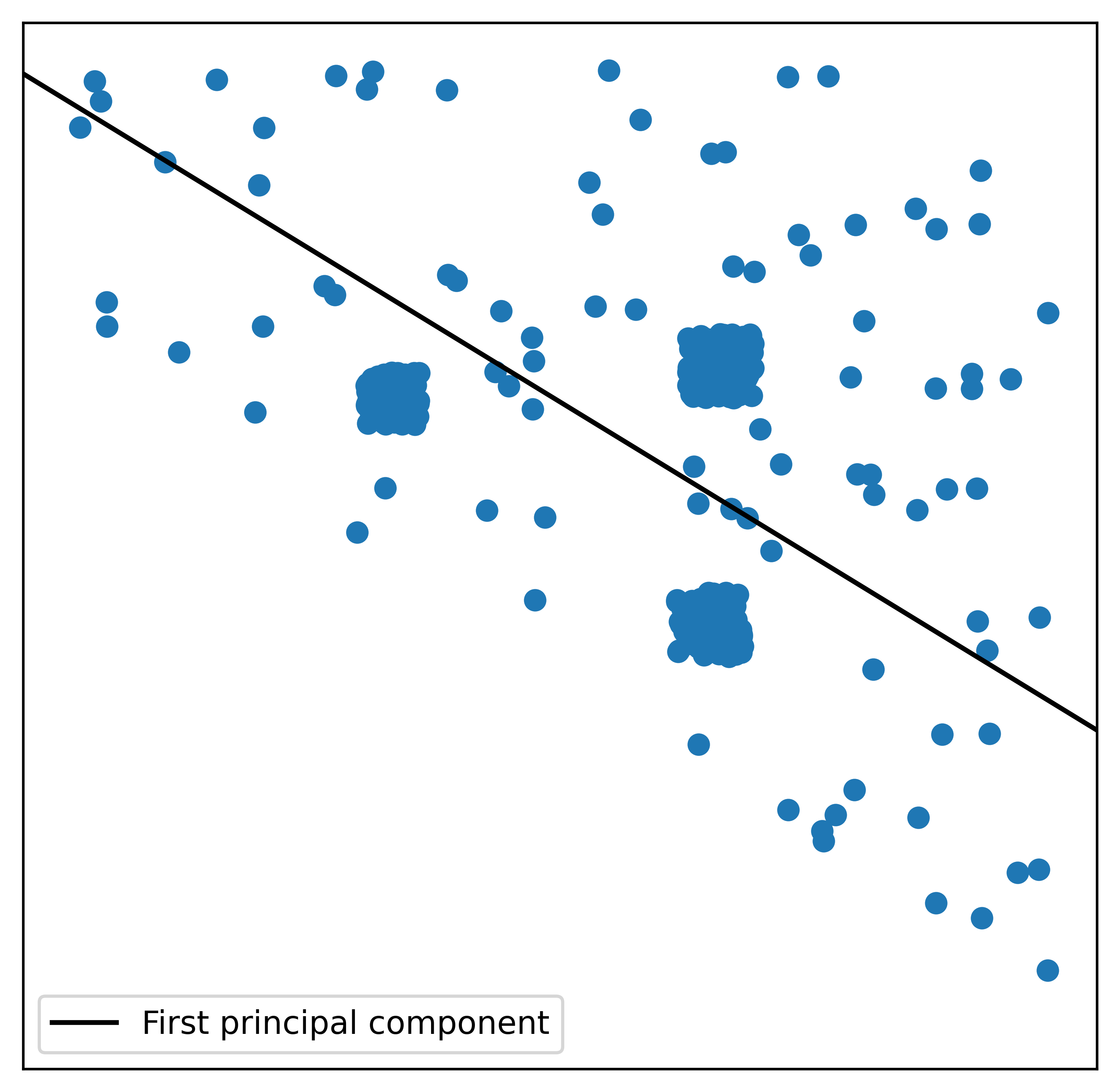}  
	\includegraphics[width=0.235\textwidth]{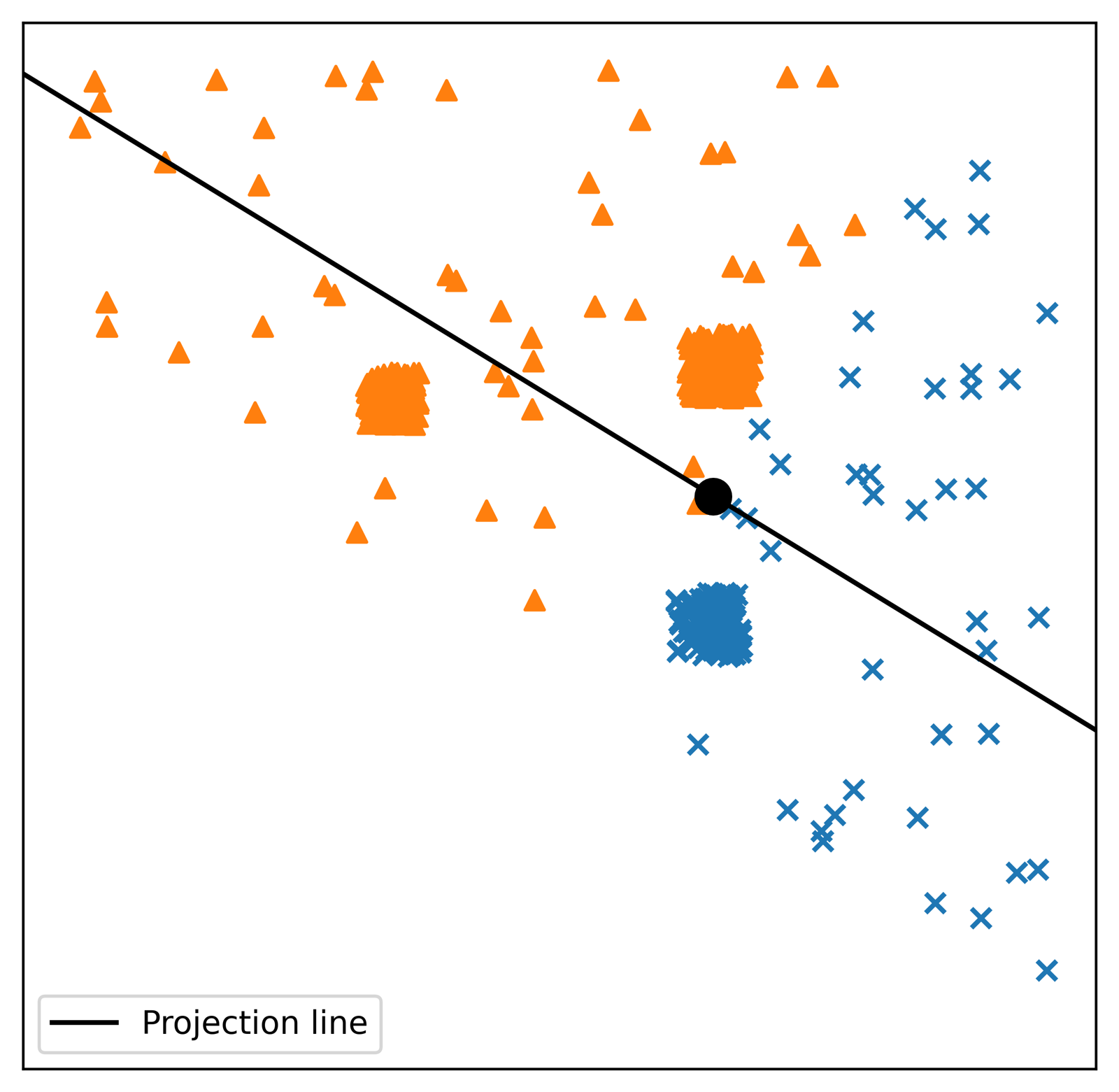}  
	\caption{\texttt{Method 4} uses PCA to find the projection dimension with the maximum dispersion; (left) the first principle component of data points in 2D; (right) splitting the points along the first principle component; the black point along the projection line is the split point. (Best viewed in color)}
	\label{Fig:Fig-05}
\end{figure}

The most optimal case when looking for the maximum dispersion of points is to run principle component analysis (PCA) \cite{Pearson1901On,abdi2010principal}. PCA guarantees to return the direction where the data points have the maximum spread. The computational complexity for PCA is $O(d^2n+d^3)$. The method that uses PCA to project data points is referred to as \texttt{Method 4} in our experiments. An example of \texttt{Method 4} is shown in Figure\ \ref{Fig:Fig-05}.

\subsection{Searching for nearest neighbors in rpTrees}
\label{SearchingrpTrees}

\begin{figure}[!t]
	\centering
	\includegraphics[width=0.45\textwidth,height=20cm,keepaspectratio]{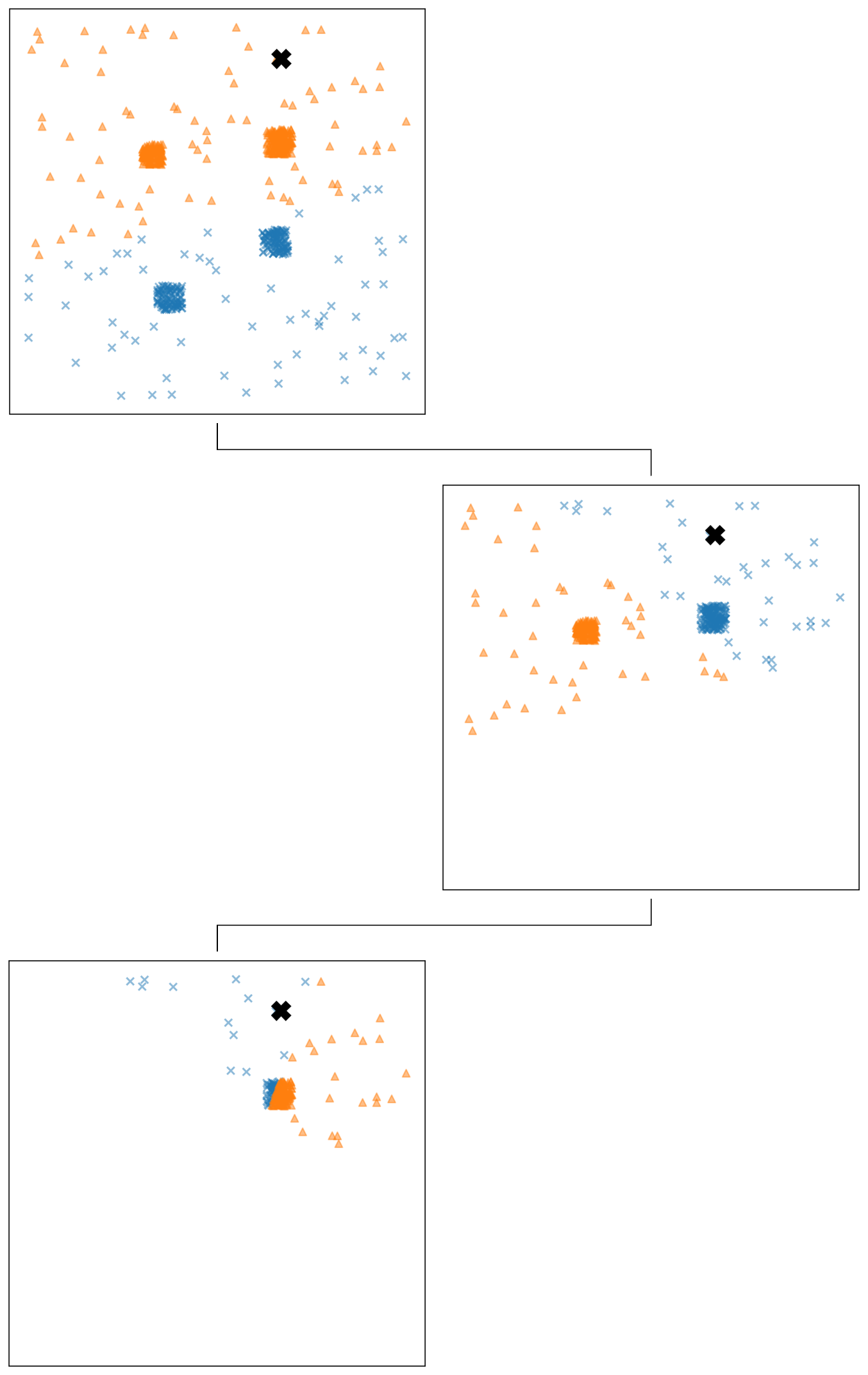}	
	\caption{$k$-nn search in rpTree; a test sample depicted by $\times$ is traversed down the tree and traced left if it sits in a blue cluster and right if otherwise. (Best viewed in color)}
	\label{Fig:Fig-06}
\end{figure}

Once all rpTrees are constructed, we are ready to perform $k$-nearest neighbor search in random projection forests (rpForests). Each rpTree node $W$ must store the projection hyperplane $\overrightarrow{r}$ used to project the data points in that node. It also has to store the split point $c$ used to split the points into the left child $W_L$ and the right child $W_R$. When searching for the $k$-nearest neighbors of a test sample $x$, we traverse $x$ from the root node all the way down to a leaf node. There are two steps involved when traversing $x$ down an rpTree: 1) use the stored projection hyperplane $\overrightarrow{r}$ to project $x$ onto it, and 2) use the stored split point $c$ to place $x$ in the left or right child. An example of walking a test sample down an rpTree is shown in Figure\ \ref{Fig:Fig-06}.

rpForest is a collection of rpTrees, each of which is constructed by one of the methods explained in the previous section \texttt{Method 1} to \texttt{Method 4}. The $k$-nearest neighbors returned from all rpTrees in the rpForest are aggregated into one set. After removing the duplicate points, the nearest neighbors are returned using Euclidean distance in the function \texttt{(scipy.spatial.distance.cdist)} \cite{2020SciPy-NMeth}. In case of a tie (i.e., two neighbors having the same distance from the test sample $x$), the rank of neighbors is decided by the function \texttt{(numpy.argsort)} \cite{harris2020array}.

To evaluate the results returned by $k$-nearest neighbor search in rpForests, we used two metrics proposed by (Yan et al., 2021) \cite{Yan2021Nearest} and (Yan et al., 2018) \cite{Yan2018Nearest}. The first metric returns a fraction representing the true neighbors $k$ missed by $k$-nn search in rpForests. The true neighbors are retrieved using a brute-force method. Let $x_1, x_2, \cdots, x_n$ be the data points in a dataset. For a point $x_i$, the number of true neighbors missed by $k$-nn search is denoted by $m_i$. The average missing rate $\overline{m}$ is defined as:

\begin{equation}
	\overline{m}=\frac{1}{nk} \sum_{i=1}^n m_i\ .
	\label{Eq-002}
\end{equation}

The second metric measures the discrepancy between the distance to the true $k^{\text{th}}$ neighbor $d_k(i)$ and the distance to the $k^{\text{th}}$ neighbor found by the algorithm $\hat{d}_k(i)$. The average distance error $\overline{d}_k$ is defined as:

\begin{equation}
	\overline{d}_k=\frac{1}{n} \sum_{i=1}^n \hat{d}_k(i) - d_k(i)\ .
	\label{Eq-003}
\end{equation}
\section{Experiments and discussions}
\label{Experiments}

\begin{figure*}
	\centering
	\includegraphics[width=0.24\textwidth]{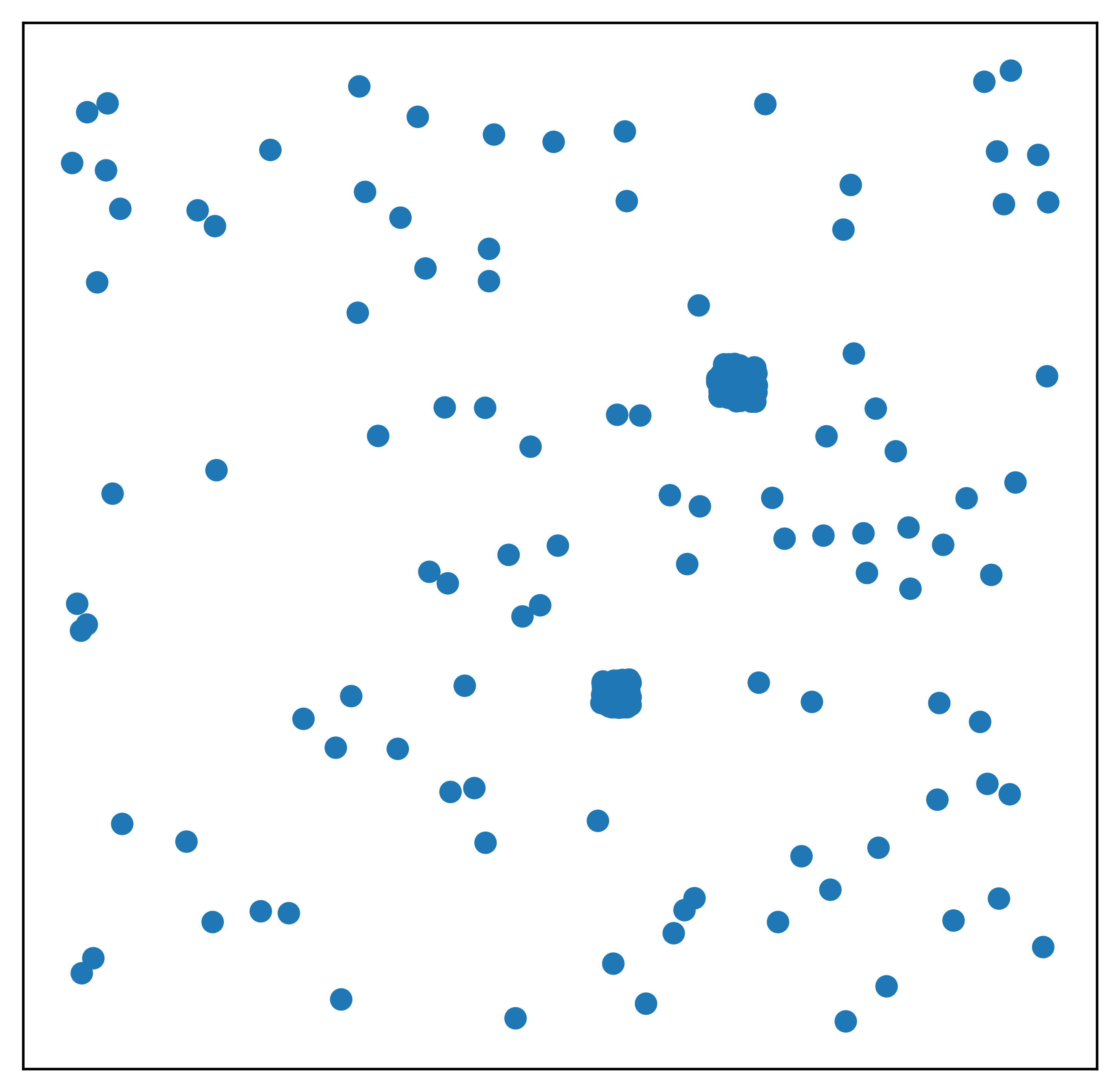}  
	\includegraphics[width=0.24\textwidth]{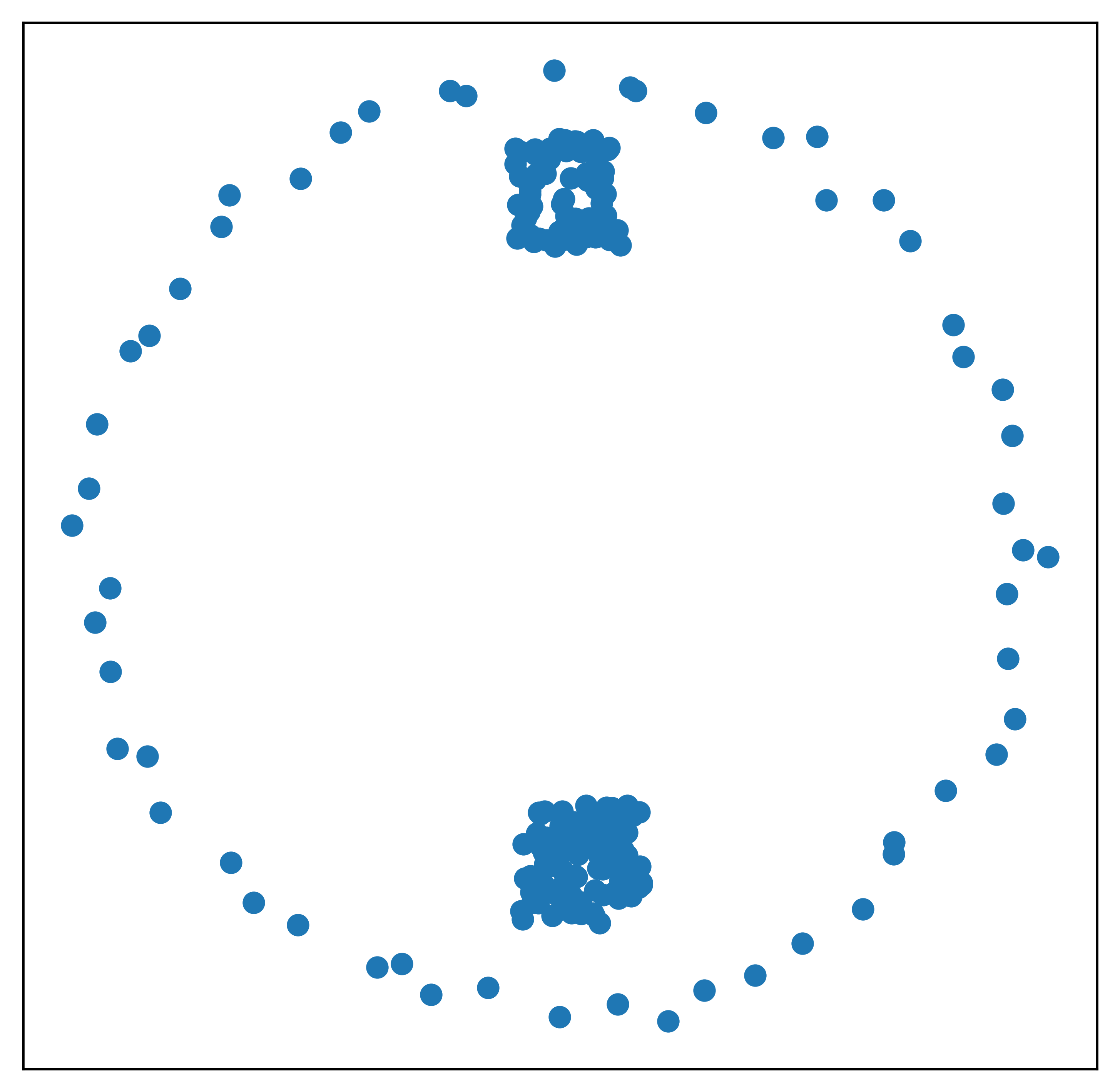}
	\includegraphics[width=0.24\textwidth]{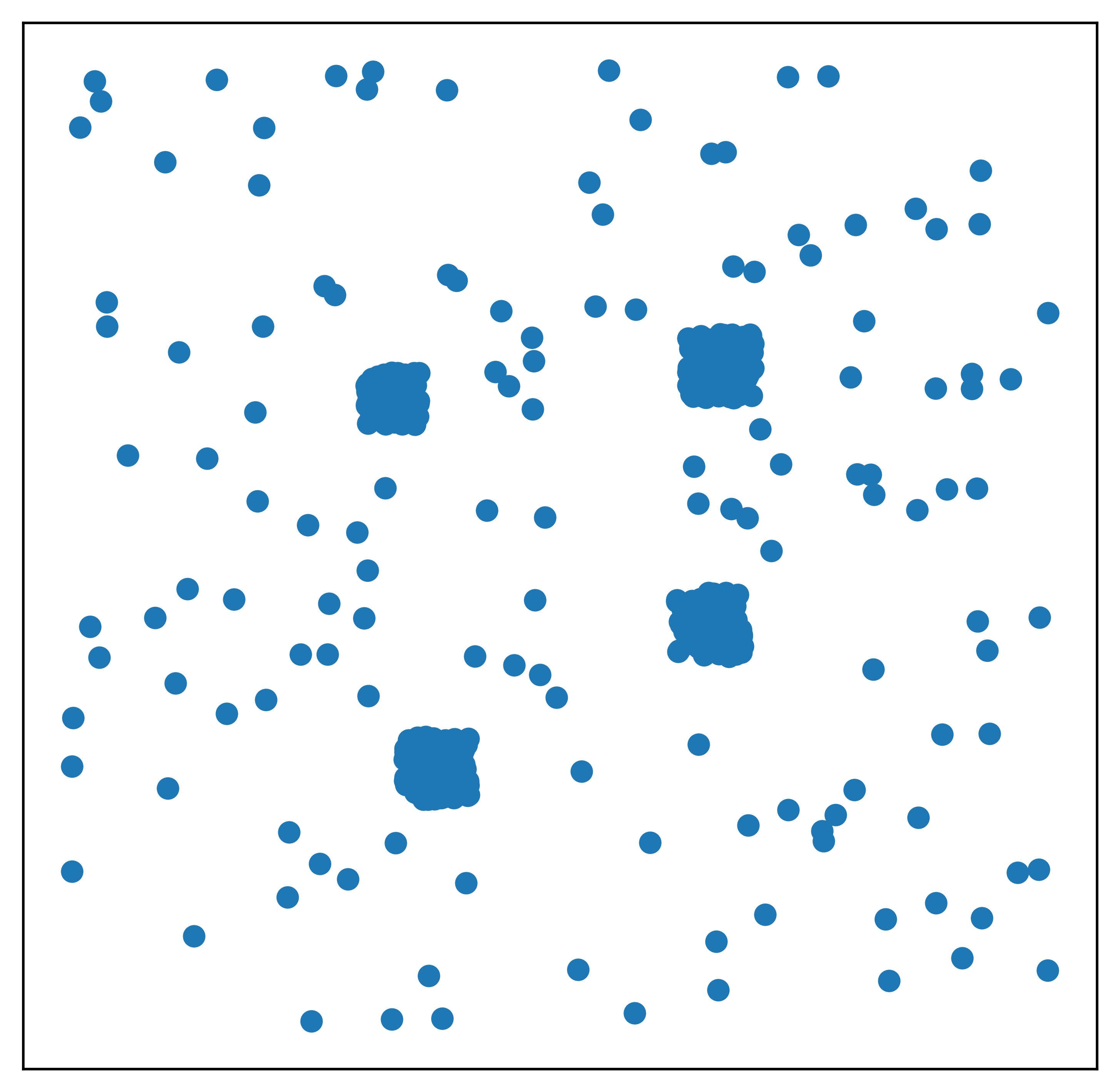}  
	\includegraphics[width=0.24\textwidth]{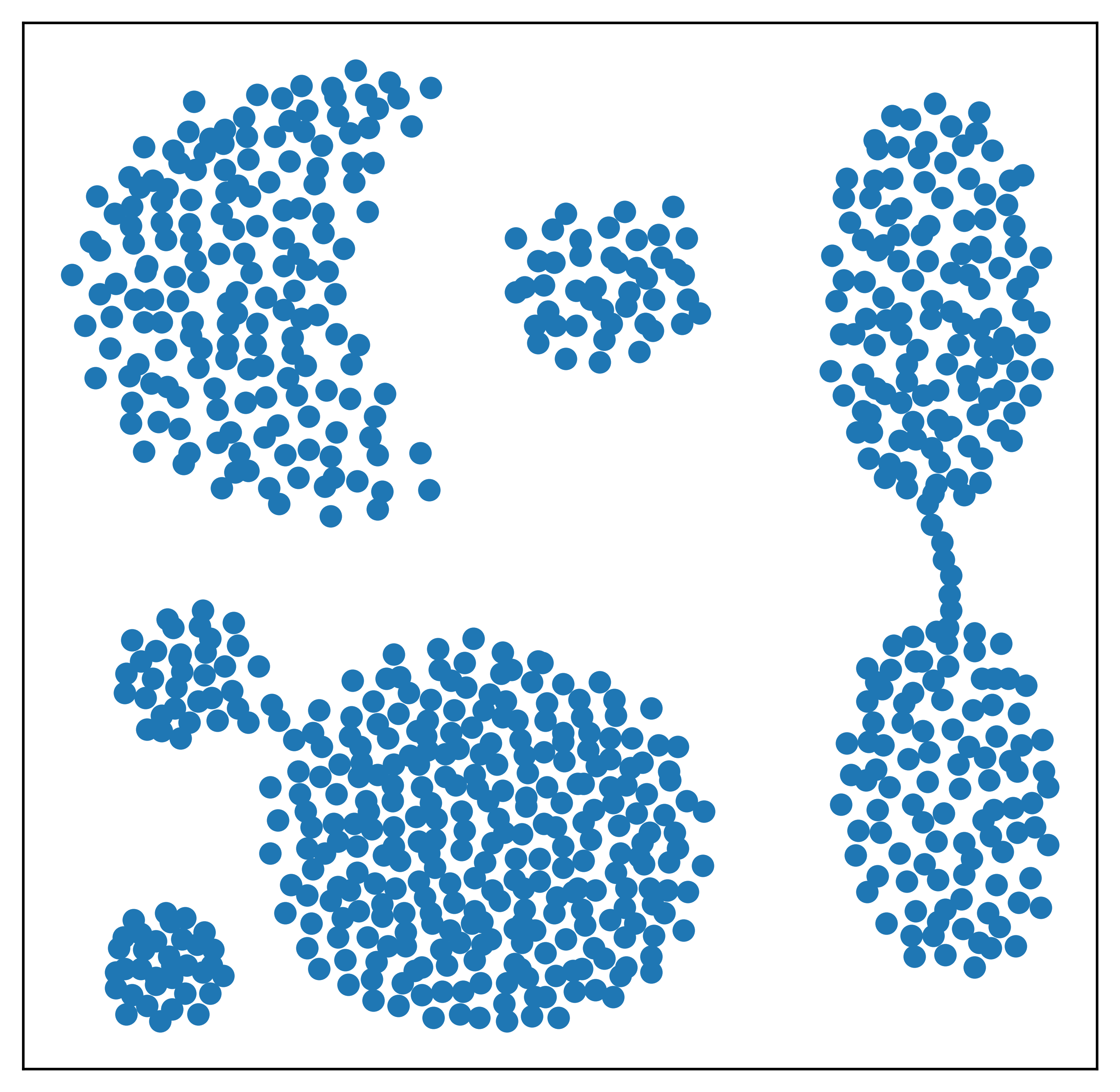}	  
	\caption{Synthetic datasets used in the experiments; from left to right \texttt{Dataset 1} to \texttt{Dataset 4}; source: \cite{Zelnik2005Self,ClusteringDatasets}.}
	\label{Fig:Fig-07}
\end{figure*}

\begin{table*}
	\centering
	\caption{Properties of tested datasets. $N =$ number of samples, $d =$ number of dimensions (i.e., features).}
	\begin{tabular}{l rrr | l rrr | l rrr} 
		\hline\hline \\ [-1.5ex] 
		\multicolumn{4}{c}{Synthetic datasets} & 
		\multicolumn{4}{c}{Small real datasets} & 
		\multicolumn{4}{c}{Large real datasets}\\ [0.5ex]
		\hline 
		& $N$ & $d$ & source &
		& $N$ & $d$ & source &
		& $N$ & $d$ & source\\ 
		
		Dataset 1	& 303 & 2 & \cite{Zelnik2005Self} &
		iris	 	& 150 & 4 & \cite{scikit-learn} &
		mGamma & 19020 & 10 & \cite{Dua2019UCI}\\
		
		Dataset 2	& 238 & 2 & \cite{Zelnik2005Self} &
		wine		& 178 & 13 & \cite{scikit-learn} &
		Cal housing & 20640 & 8 & \cite{scikit-learn}\\		
		
		Dataset 3	& 622 & 2 & \cite{Zelnik2005Self }&
		BC-Wisc.	& 569 & 30 & \cite{scikit-learn} &
		credit card & 30000 & 24 & \cite{Dua2019UCI}\\
		
		Dataset 4	& 788 & 2 & \cite{ClusteringDatasets} &
		digits		& 1797 & 64 & \cite{scikit-learn} &
		CASP 		& 45730 & 9 & \cite{Dua2019UCI}\\		
		\hline 
	\end{tabular}
	\label{Table:Table-01}
\end{table*}

\begin{figure*}
	\centering
	\begin{tabular}{c | c } 
		\hline \\ [-1.5ex]
		$T \in \{1, 2, 3, 4, 5\}$ &
		$T \in \{10, 20, 40, 60, 80, 100\}$ \\ [0.5ex]
		\hline 
		\raisebox{-\totalheight}{\includegraphics[width=0.49\textwidth]{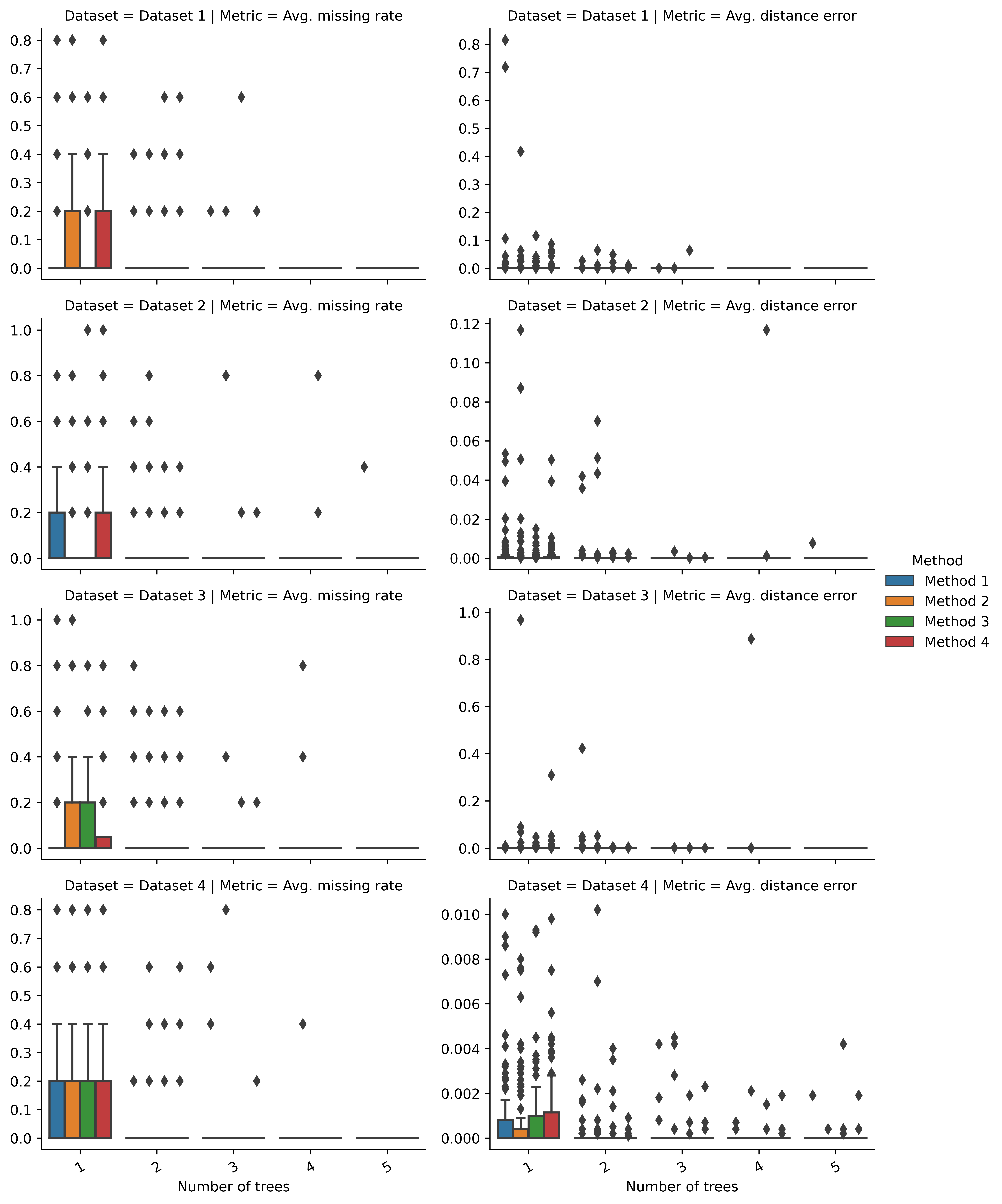}} &
		\raisebox{-\totalheight}{\includegraphics[width=0.49\textwidth]{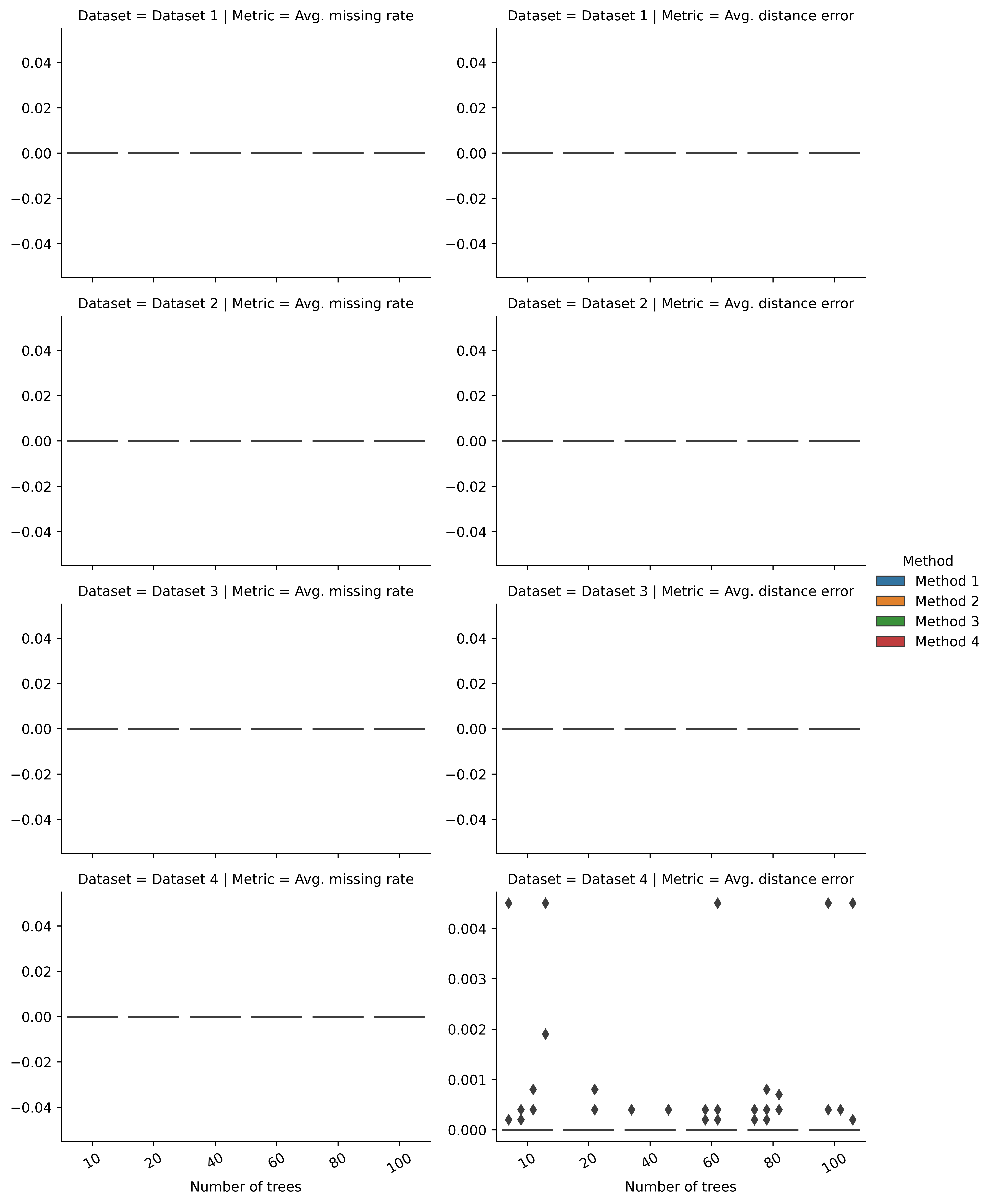}} \\				
		\hline 
	\end{tabular}
	\caption{Average missing rate $\overline{m}$ and average distance error $\overline{d}_k$ after testing the four methods on four synthetic datasets: \texttt{Dataset 1, Dataset 2, Dataset 3 } and \texttt{, Dataset 4}. Each box represents a 100 runs average. (Best viewed in color)}
	\label{Fig:Fig-synthetic}
\end{figure*}

\begin{figure*}
	\centering
	\begin{tabular}{c | c } 
		\hline \\ [-1.5ex]
		$T \in \{1, 2, 3, 4, 5\}$ &
		$T \in \{10, 20, 40, 60, 80, 100\}$ \\ [0.5ex]
		\hline 
		\raisebox{-\totalheight}{\includegraphics[width=0.49\textwidth]{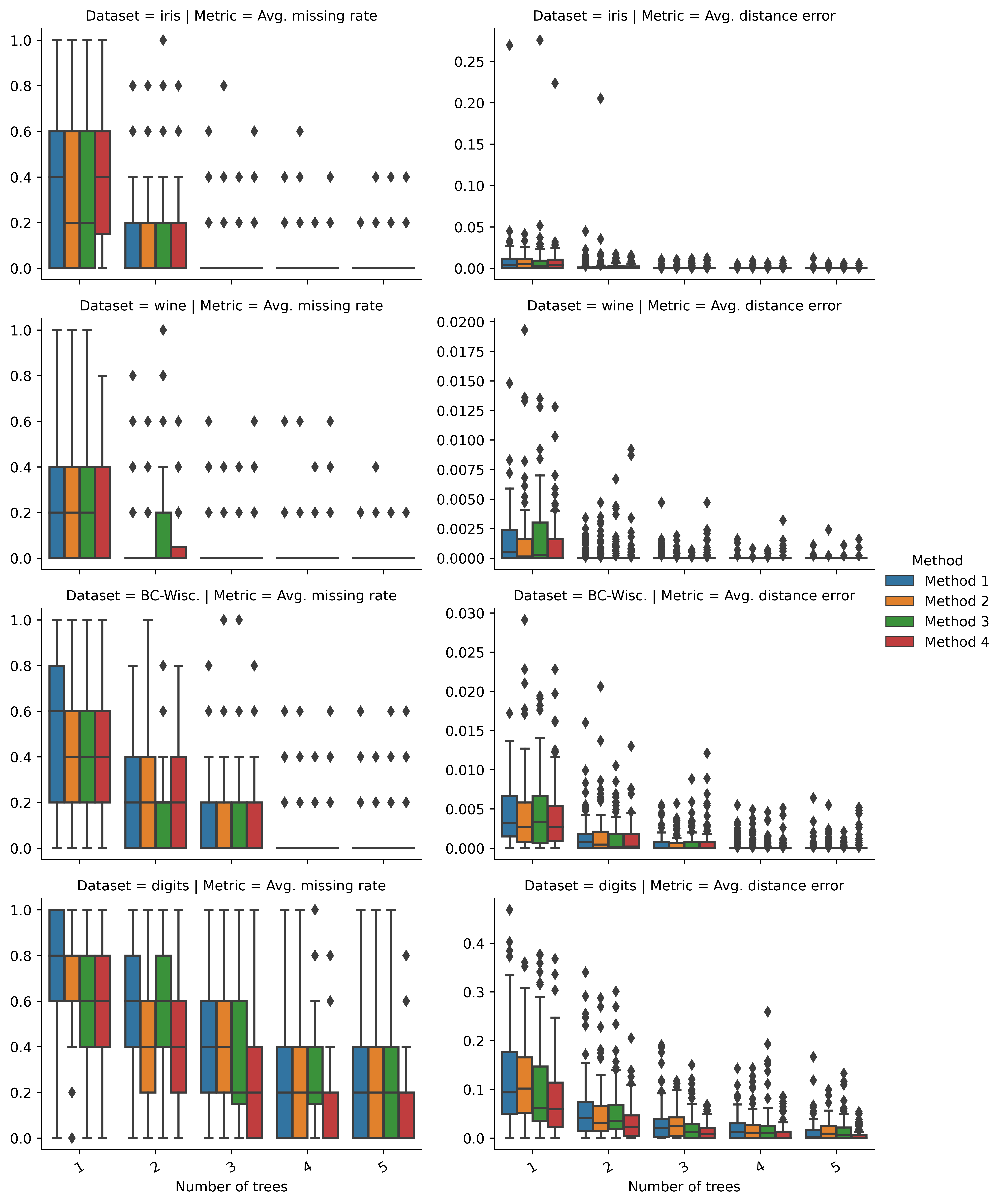}} &
		\raisebox{-\totalheight}{\includegraphics[width=0.49\textwidth]{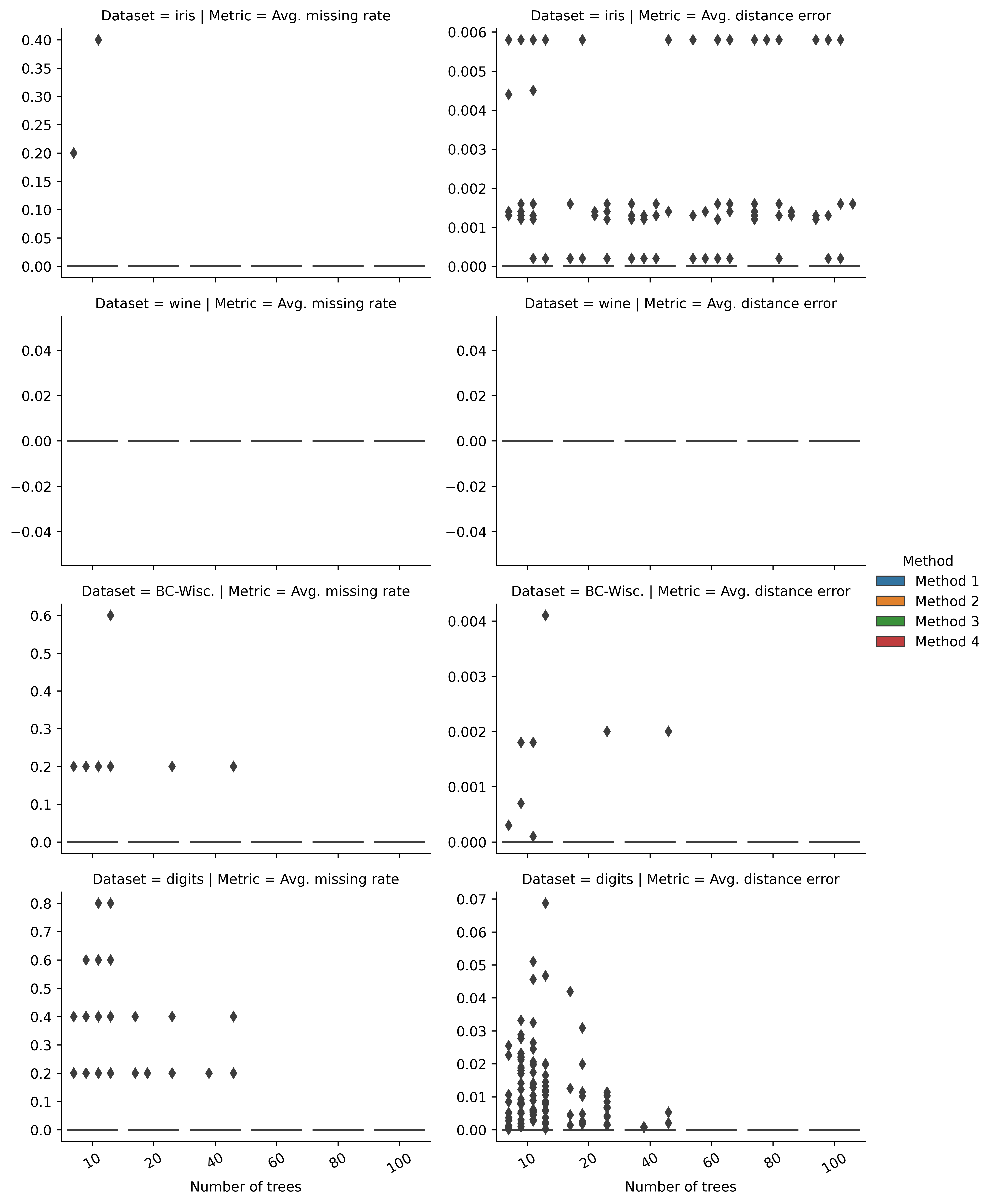}} \\				
		\hline 
	\end{tabular}
	\caption{Average missing rate $\overline{m}$ and average distance error $\overline{d}_k$ after testing the four methods on four small real datasets: \texttt{iris, wine, BC-Wisc. } and \texttt{, digits}. Each box represents a 100 runs average. (Best viewed in color)}
	\label{Fig:Fig-small}
\end{figure*}

\begin{figure*}
	\centering
	\begin{tabular}{c | c } 
		\hline \\ [-1.5ex]
		$T \in \{1, 2, 3, 4, 5\}$ &
		$T \in \{10, 20, 40, 60, 80, 100\}$ \\ [0.5ex]
		\hline 
		\raisebox{-\totalheight}{\includegraphics[width=0.49\textwidth]{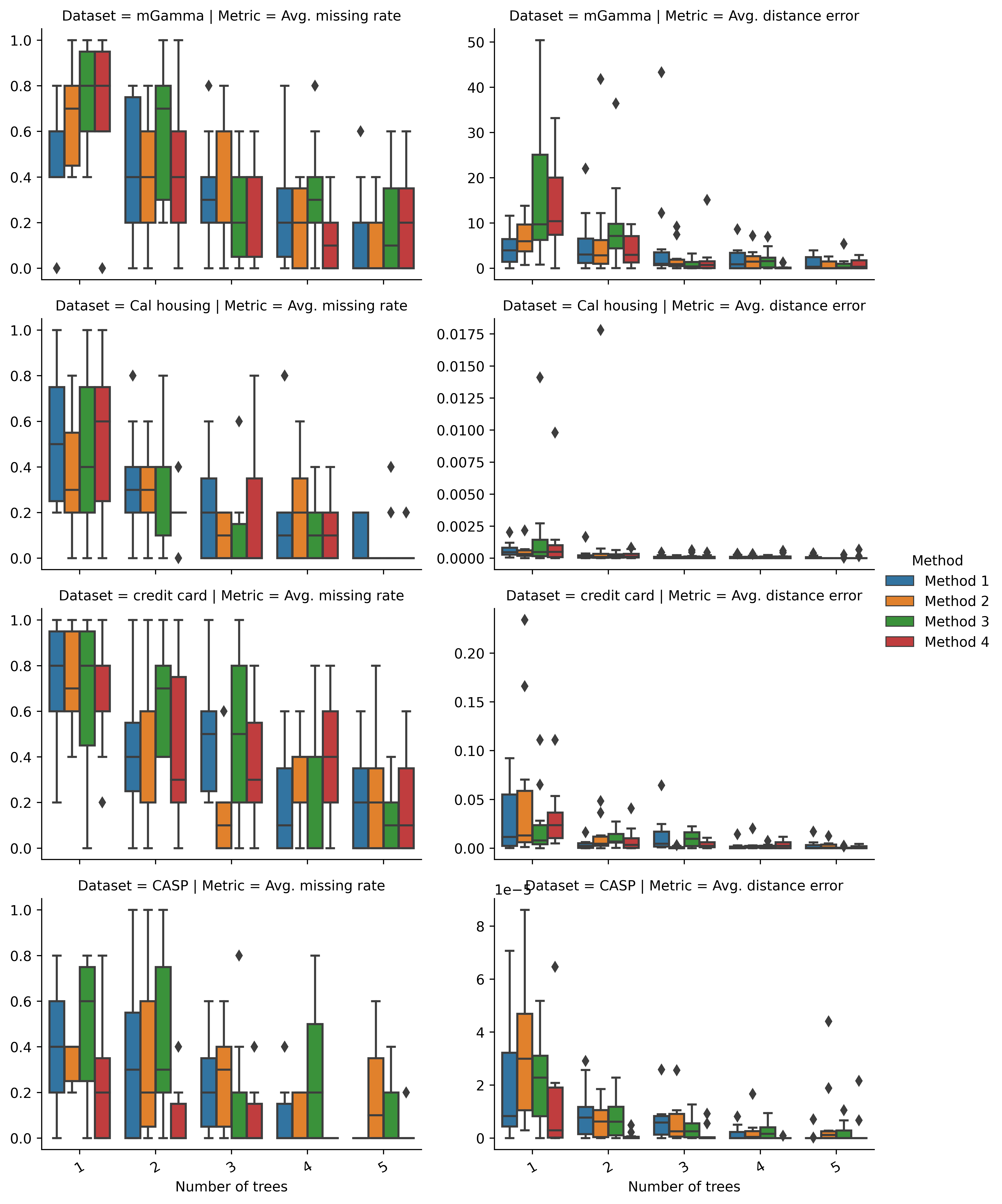}} &
		\raisebox{-\totalheight}{\includegraphics[width=0.49\textwidth]{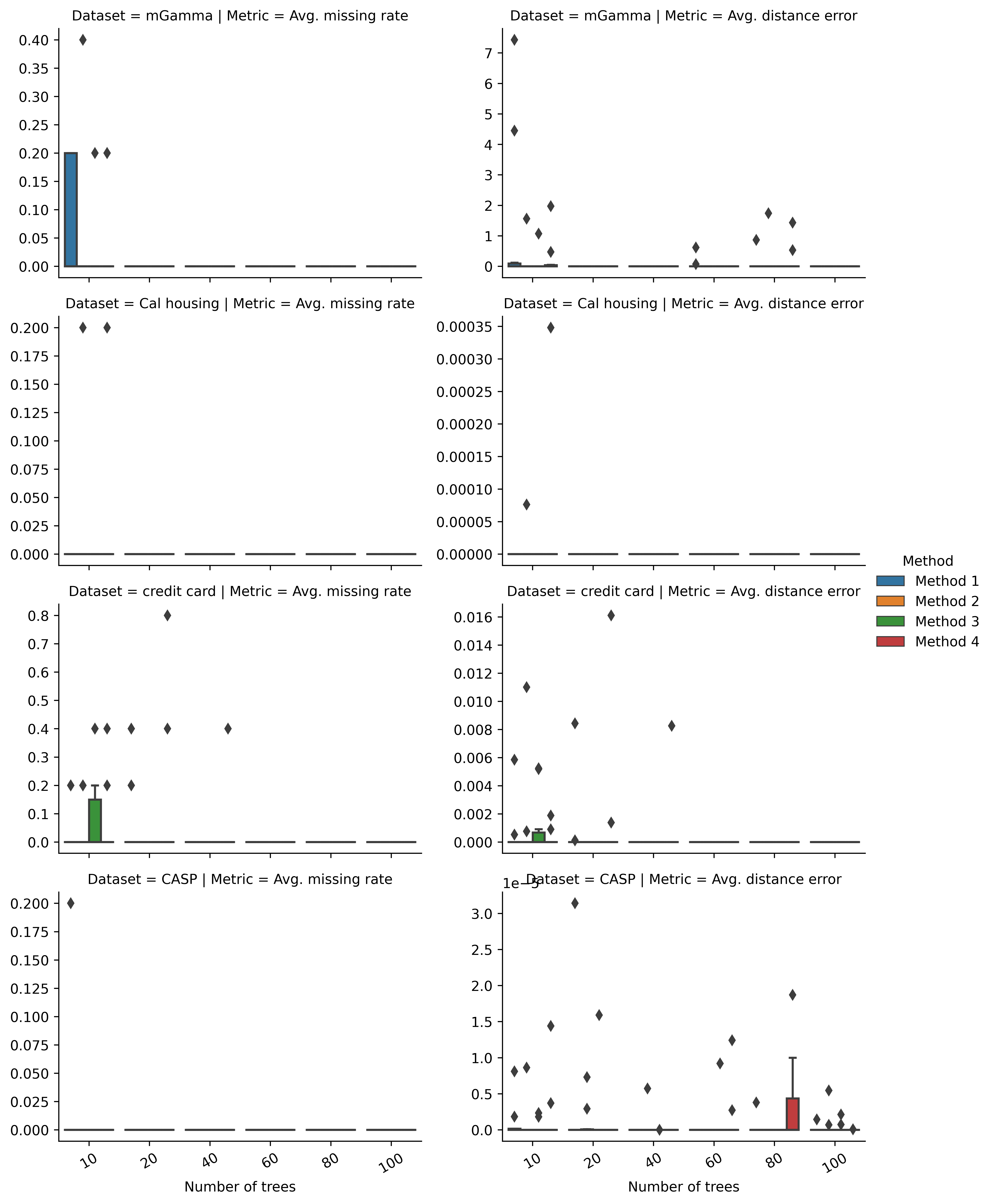}} \\				
		\hline 
	\end{tabular}
	\caption{Average missing rate $\overline{m}$ and average distance error $\overline{d}_k$ after testing the four methods on four large real datasets: \texttt{mGamma, Cal housing, credit card } and \texttt{, CASP}. Each box represents a 10 runs average. (Best viewed in color)}
	\label{Fig:Fig-Large}
\end{figure*}

The experiments were designed to investigate the relationship between the dispersion of points ($\sigma_r$) when splitting tree nodes and the number of trees ($T$). There are a couple of factors that could affect the $k$-nearest neighbor search in rpForests. The first factor is the dataset used, which includes the number of samples $n$ and the number of dimensions $d$. The second factor that affects $k$-nn search is the value of $k$, which indicates how many neighbors should be returned by the algorithm. To draw a consistent conclusion on the relationship between the dispersion of points and the number of trees, we have to test the method on datasets from different applications. We also have to repeat the experiments using different values of $k$.

The evaluation metrics used in our experiments are the ones proposed by Yan, et al. \cite{Yan2021Nearest,Yan2018Nearest}. The first metric in equation \ref{Eq-002} measures the number of neighbors missed by the algorithm. The second metric (shown in equation \ref{Eq-003}) measures the distance to the $k^{\text{th}}$ neighbor found by the algorithm. Apparently, if the algorithm made a mistake, the distance to the $k^{\text{th}}$ neighbor will be larger than the distance to the true $k^{\text{th}}$ neighbor.

The code used to produce the experiments is available on \url{https://github.com/mashaan14/RPTree}. To make our experiments reproducible, we used standardized datasets and off the shelf machine learning libraries. The properties of the tested datasets are shown in Table\ \ref{Table:Table-01}. The functions to perform random projection and principle component analysis (PCA) in our experiments are the ones provided by scikit-learn library in python \cite{scikit-learn, sklearn_api}. All experiments were coded in python 3 and run on a machine with 20 GB of memory and a 3.10 GHz Intel Core i5-10500 CPU.

\subsection{Experiments on synthetic and real data}
\label{SyntheticRealData}

In the first part of this experiment, we used four synthetic datasets with different distribution of data points. The value of $k$ was set to 5 and the rpTree node $W$ was not allowed to have less than 20 data points. The results of this experiment are shown in Figure\ \ref{Fig:Fig-synthetic}. An observation that kept persisting across all datasets is that all methods performed better with a large number of rpTrees. The missing rate remained low for all methods when $T>20$.  Also, the distance error remained low when $T>20$.

\texttt{Method 1} has the highest missing rate when $T=1$ in three of the four datasets. This is due to the randomness involved in this method, since it does not maximize the dispersion of points before splitting the points. This increases the chance that two true neighbors are placed in different tree branches. When we increased the number of trees $T>20$, \texttt{Method 1} delivered similar performance as the other competing methods as shown in the right side of Figure\ \ref{Fig:Fig-synthetic}. Since \texttt{Method 1} involves less computations than other methods, it is recommended to use it when $T>20$.

The second part of this experiment uses small real datasets shown in Table\ \ref{Table:Table-01}. The same settings in synthetic datasets were used here. $k=5$ and the capacity of the node $W$ should not drop below 20 data points. The results of this experiment are shown in Figure\ \ref{Fig:Fig-small}, where all boxes are averages of 100 runs.

We continue to have the same observation from the last experiment. When the number of trees is large $T>20$, it does not matter which method you used to construct rpTrees. So, the user is better off using the original method \texttt{Method 1} because it has less computations.

We also noticed that in the last dataset, the \texttt{digits} dataset, the methods needed large number of trees usually above 20, to maintain a lower missing rate as shown in the right side of Figure\ \ref{Fig:Fig-small}. This was not the case in the previous three datasets \texttt{iris}, \texttt{wine}, and \texttt{breast cancer}, when the missing rate dropped and stabilized well before we reach $T=20$. This can be explained by the large number of data points in \texttt{digits} dataset compared with the other three datasets.

The last part of this experiment was to test the competing methods on large real datasets: \texttt{mGamma, Cal housing, credit card } and \texttt{, CASP}. The properties of these datasets are shown in Table\ \ref{Table:Table-01}. The results of this experiment are shown in Figure\ \ref{Fig:Fig-Large}, in which all boxes represent an average of 10 runs.

With \texttt{mGamma} dataset, the $k$-nn search was ineffective when $T=1$, regardless of the method used. The competing methods have some runs where the average missing rate $\overline{m}=1$, indicating the methods could not retrieve any of the true neighbors when $T=1$. As we increase $1 < T \le 5$, we notice that $\overline{m}$ dropped below $0.2$, which means the methods were able to retrieve some of the true neighbors. With $T > 20$ $\overline{m}$ and $\overline{d}_k$ stabilize around 0, which means the methods are able to retrieve all true neighbors, and the distance to the $k^{\text{th}}$ neighbor found by the methods equals the true distance.

The observation we had with \texttt{mGamma} dataset when increasing the value of $T$ applies to other datasets in Figure\ \ref{Fig:Fig-Large} \texttt{Cal housing, credit card } and \texttt{, CASP}. We also observed that some runs with \texttt{credit card} dataset had high $\overline{m}$ value even with large number of trees with $20 < T \le 40$. This can be explained by the high dimensions in this dataset $d=24$, which affects the $k$-nn search. But with $T > 60$, the average missing rate $\overline{m}$ stabilizes around zero for all competing methods.

\begin{table*}
	\centering
	\caption{Results of running t-test on the samples collected from different methods using large real datasets: \texttt{mGamma, Cal housing, credit card } and \texttt{, CASP}. $T$ is the number of trees; (-) indicates that the samples have identical averages.}
	\includegraphics[width=\textwidth,height=20cm,keepaspectratio]{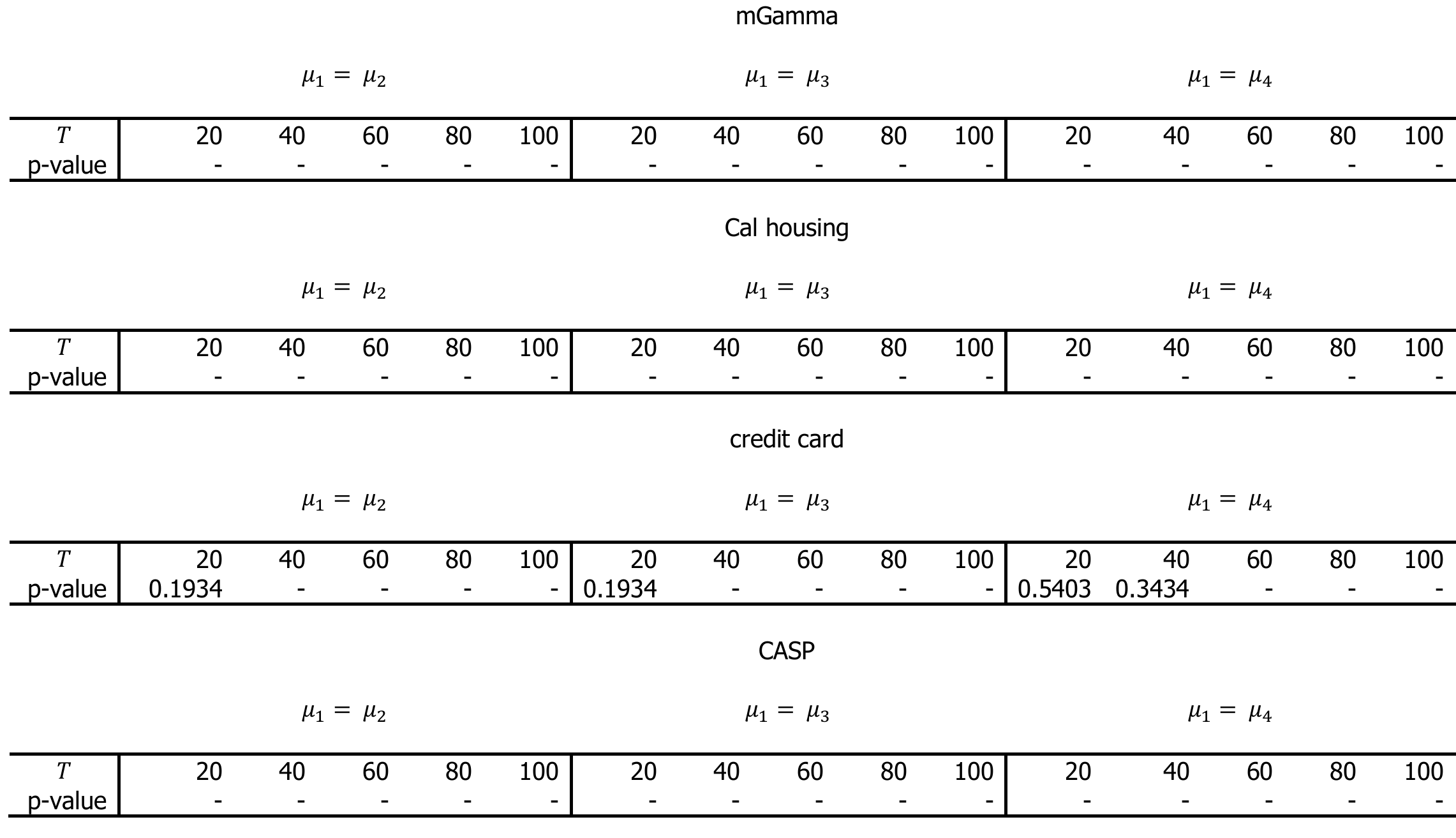}	
	\label{Table:Hypothesis-testing}
\end{table*}

\subsection{Hypothesis testing}
\label{Hypothesis-testing}

We performed a statistical significance test to evaluate the samples we collected from our experiment. The hypothesis testing was formulated as follows:

\begin{itemize}
	\item $H_0$: $\mu_A = \mu_B$ (maximizing the points dispersion does not change k-nn search results with $T>20$)
	
	\item $H_1$: $\mu_A \ne \mu_B$ (maximizing the points dispersion changes k-nn search results with $T>20$)
\end{itemize}
\noindent
$\mu_A$ is the average missing rate $\overline{m}$ using \texttt{Method 1}, and $\mu_B$ is the average missing rate $\overline{m}$ using \texttt{Method 2}, \texttt{Method 3}, or \texttt{Method 4}. 

We used the function (\texttt{scipy.stats.ttest\_ind}) \cite{2020SciPy-NMeth} to perform t-test. The results are shown in Table\ \ref{Table:Hypothesis-testing}. By looking at Table\ \ref{Table:Hypothesis-testing} we can see that in most cases the collected samples have identical averages, which was indicated by a dash (-). In some cases, particularly in \texttt{credit card} dataset, the averages were not identical. However, the p-value was greater than $0.05$, which is not statistically significant. Based on the results shown in Table\ \ref{Table:Hypothesis-testing}, we fail to reject the null hypothesis $H_0$.

\subsection{Experiments using a varying value of $k$}
\label{Varyk}

\begin{figure*}
	\centering  
	\includegraphics[width=0.7\textwidth]{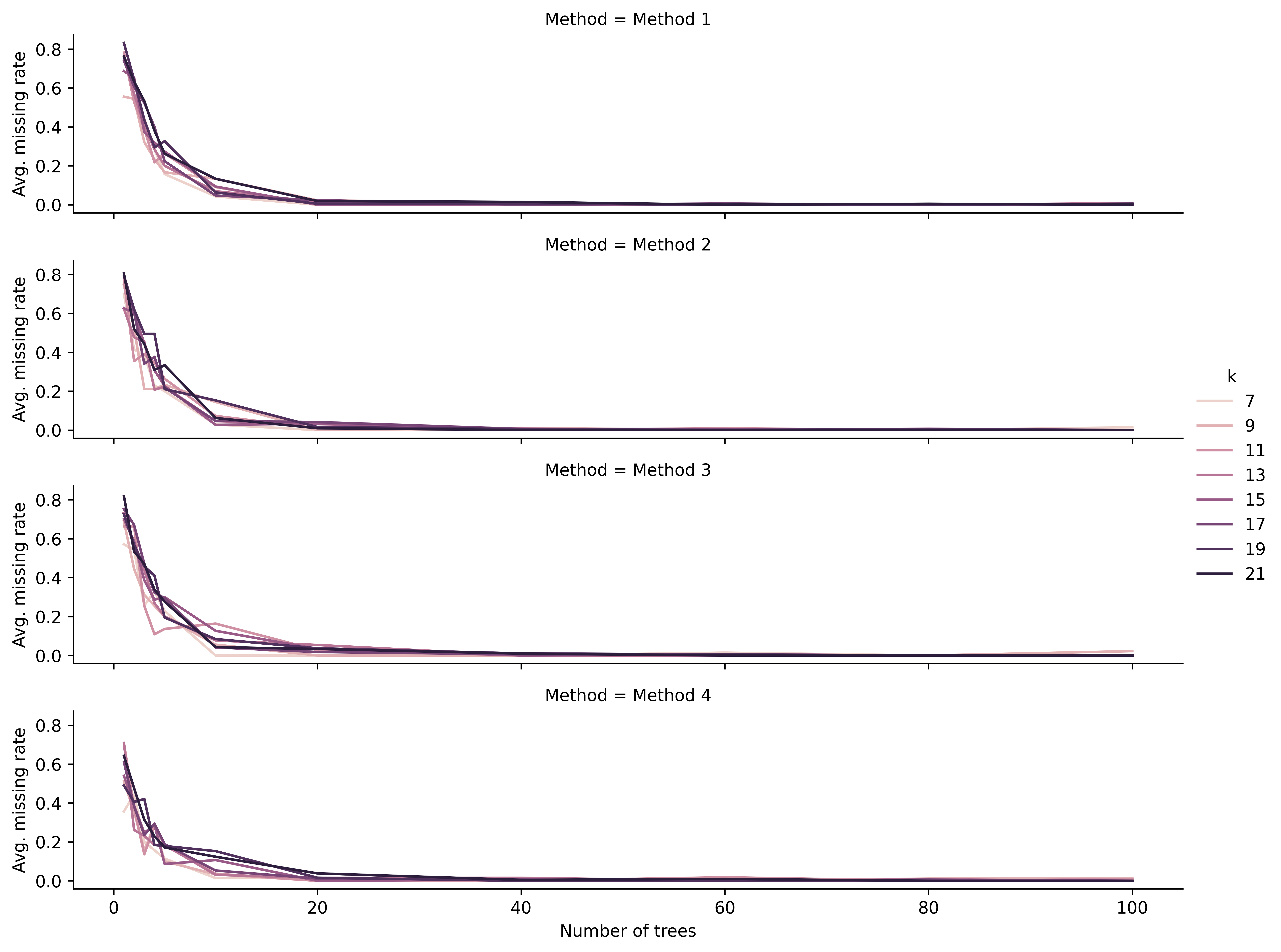}	  
	\caption{Average missing rate $\overline{m}$ after testing different values of $k$ on the \texttt{digits} dataset. Each point in the plot represents a 10 runs average. (Best viewed in color)}
	\label{Fig:Fig-12}
\end{figure*}

In this experiment we fixed the dataset factor and vary the value of the returned nearest neighbors (i.e., $k$). In the previous experiment the value of $k$ was fixed to be $5$, and the number of data points in an rpTree node $W$ should not be less than $20$. This is the same settings used by Yan, et al. \cite{Yan2021Nearest}. With a varying number of $k$ we had to increase the capacity of the node $W$ to be 30 instead of $20$. This was sufficient to cover the tested $k$ values, which are $k \in \{7,9,11,13,15,17,19,21\}$. The results are shown in Figure\ \ref{Fig:Fig-12}.

In general, with an increasing number of rpTrees $T$ in rpForest, it does not matter which method for dispersion of points you used. We also observed that with smaller values of $k$ (roughly $k < 11$), the missing rate drops faster as we increase the number of trees $T$. This can be explained by the capacity of the node $W$. With larger $k$ values we are approaching the capacity of $W$, which was set to 30. This increases the chance of missing some neighbors, as the method requires all $k$ nearest neighbors have to be present at the current node $W$. This is not the case when $k$ is small, because these few $k$ nearest neighbor have a higher chance of ending up in the same node $W$. Consequently, we need a few trees for small $k$ values, and sufficiently large number of trees (ideally $T > 20$) for large $k$ values.

\begin{figure*}
	\centering  
	\includegraphics[width=0.7\textwidth]{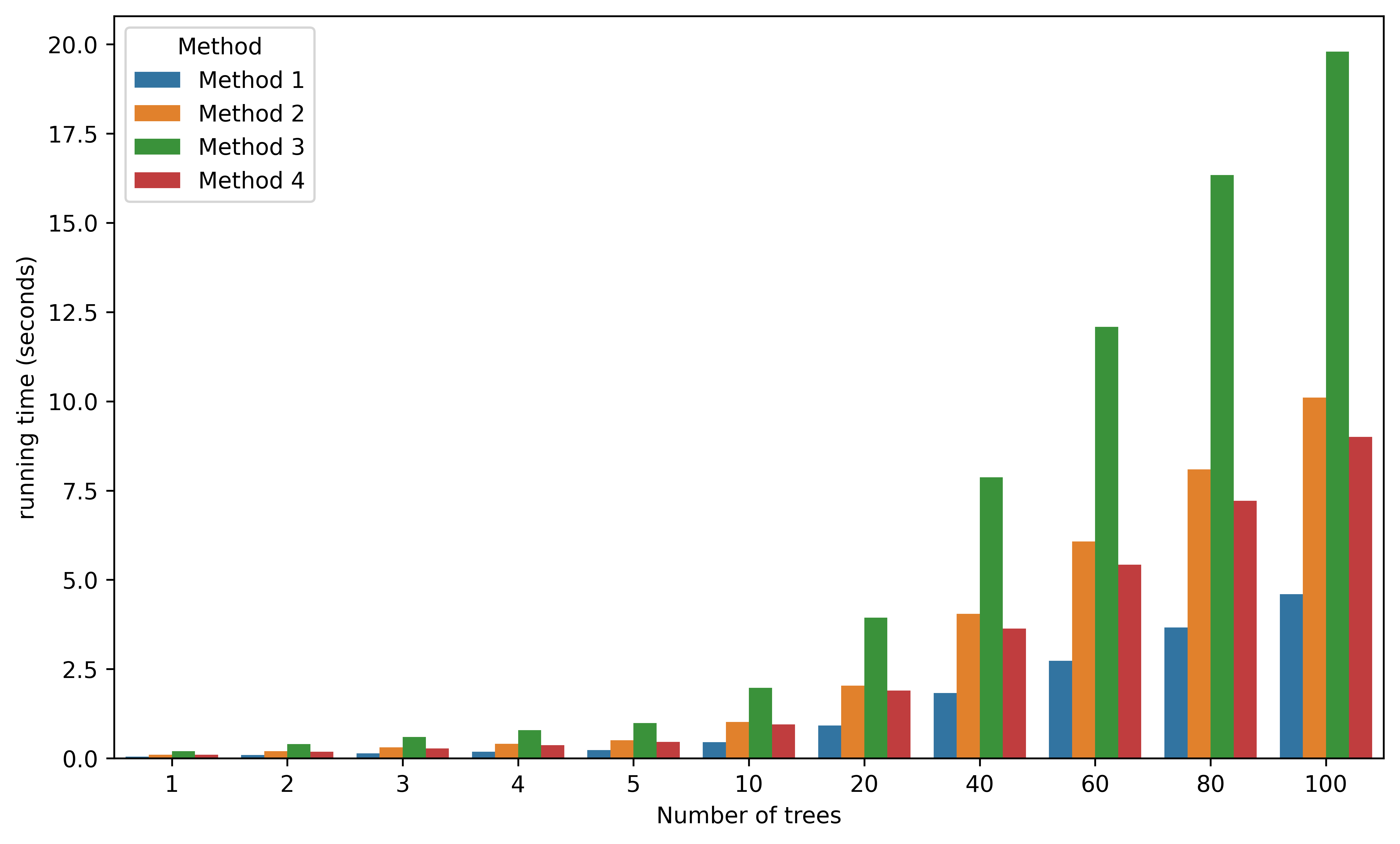}	  
	\caption{Average running time after testing the four methods on the \texttt{digits} dataset. Each bar represents a 100 runs average. We used a machine with 20 GB of memory and a 3.10 GHz Intel Core i5-10500 CPU. (Best viewed in color)}
	\label{Fig:Fig-13}
\end{figure*}

\subsection{Experiments on running time}
\label{RunningTime}

The running time experiment (shown in Figure\ \ref{Fig:Fig-13}) measures the time taken by each method to: 1) construct the rpForest, and 2) return the $k$ nearest neighbors. The value of $k$ was set to 5, and the minimum capacity of an rpTree node $W$ was set to 20 data points. The dataset used in this experiment was \texttt{digits} dataset with 1797 samples and 64 dimensions.

\texttt{Method 1} which is the original algorithm of rpTree \cite{Dasgupta2008Random} was the fastest across all experiments. This was not a surprise, since this method picks only one random direction and projects the data points onto it. On the contrary, \texttt{Method 2} and \texttt{Method 3} attempts to maximize the dispersion of points using expensive computations before splitting these points into different tree branches. \texttt{Method 4} tries to do the same using the expensive process of principle component analysis (PCA). 

A final remark on this experiment is why \texttt{Method 4} is slightly faster than \texttt{Method 2} and substantially faster than \texttt{Method 3}. This could be explained by the for loops involved in these two methods (\texttt{Method 2} and \texttt{Method 3}). \texttt{Method 2} uses a single for loop that runs for $nTry$ times to find the direction with the maximum dispersion. \texttt{Method 3} uses three for loops each of which run $nTry$ times to find the direction with the maximum dispersion and then adds noise to it to maximize the dispersion. Meanwhile, \texttt{Method 4} uses the native PCA implementation in scikit-learn, which uses an optimized library to find the principle components.

If we couple the findings from this experiment with the findings we had from the previous two experiments, we are confident to recommend the use of the original algorithm of rpTree \cite{Dasgupta2008Random}. It is true that this method might produce a random direction that could split two true neighbors. But with an increasing number of trees ($T>20$) this risk will be mitigated. Additional trees minimizes the risk of splitting two true neighbors. Using 100 rpTrees, \texttt{Method 1} is faster than \texttt{Method 2}, \texttt{Method 3}, and \texttt{Method 4} by 100\%, 300\%, and 100\% respectively. This performance gain comes with similar $k$-nearest neighbor missing rate, making \texttt{Method 1} as the recommended method for $k$-nn search in rpForests.

\section{Conclusion}
\label{Conclusion}

Random projection trees (rpTrees) generalize the concept of $k$d-trees by using random directions to split, instead of restricting the splits to the existing dimensions. An rpForest is a collection of rpTrees. The intuition is if one tree separates two true neighbors, this is less likely to happen in other rpTrees. Previous studies suggested using a random direction with the maximum dispersion to get better splits. However, it was not clear how the dispersion of points will affect the $k$-nn search in a large rpForest. We conducted experiments to investigate if the dispersion of points affects $k$-nn search in rpForest with varying number of trees. We are confident to recommend that if the number of trees is sufficient to achieve a small error rate, one is better off using random directions regardless of the dispersion of points. This method is faster than other methods with the same $k$-nn search efficiency in case of a sufficient number of rpTrees.

A possible extension of this work is to test how the examined method will perform in problems other than $k$-nn search, like spectral clustering. Also, we would like to explore if there are metrics other than the dispersion of points that improve $k$-nn search with less computational footprint.

\bibliographystyle{IEEEtran}
\bibliography{mybibfile}

\begin{IEEEbiography}[{\includegraphics[width=1in,height=1.25in,clip,keepaspectratio]{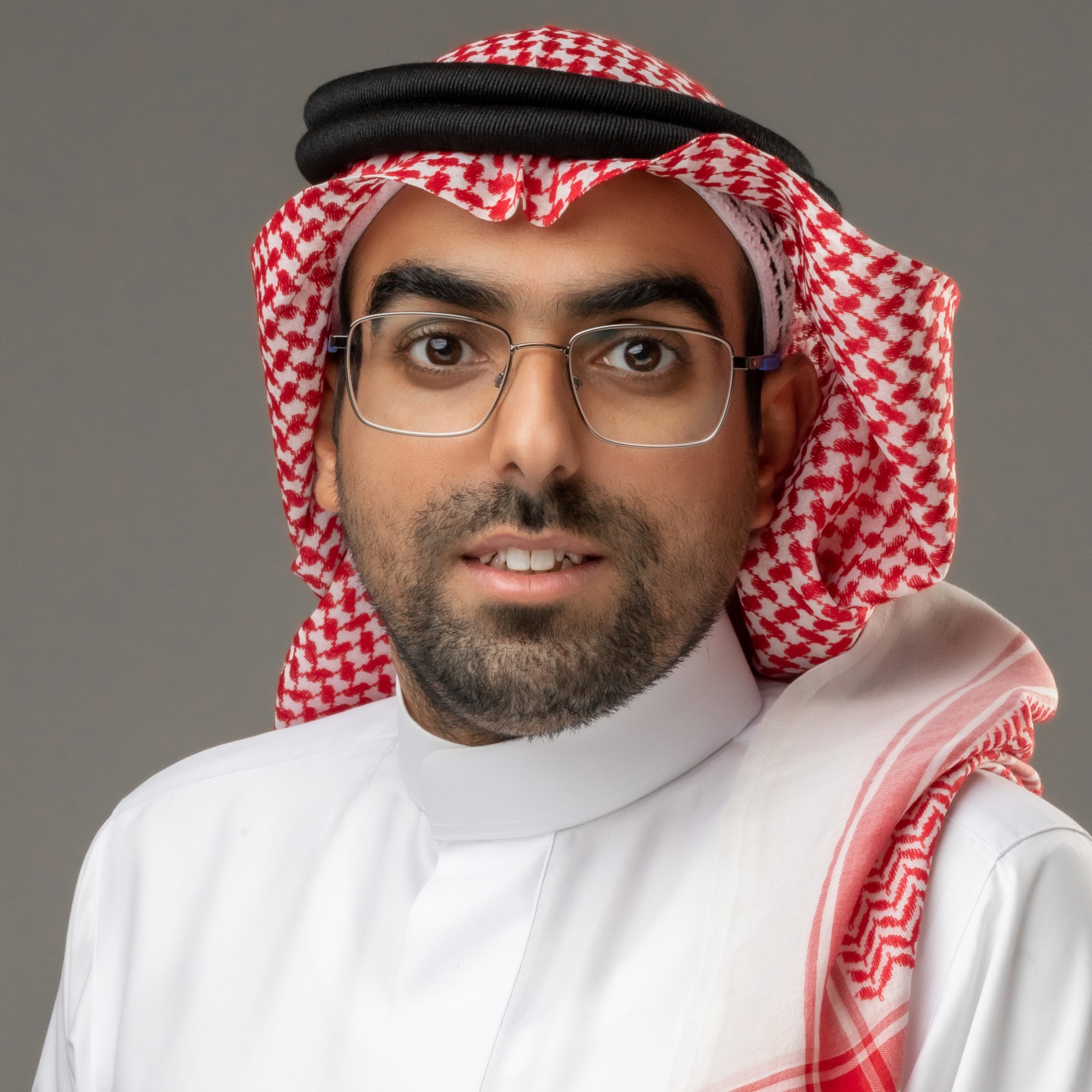}}]{Mashaan Alshammari} Dr. Mashaan holds a MSc in computer science from King Fahd University of Petroleum and Minerals (KFUPM), Saudi Arabia, and a PhD from the University of Sydney, Australia. His research interests lie broadly in graph clustering and deep representation learning.
\end{IEEEbiography}

\begin{IEEEbiography}[{\includegraphics[width=1in,height=1.25in,clip,keepaspectratio]{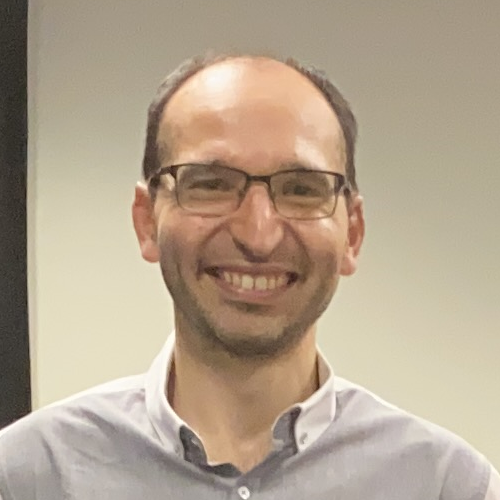}}]{John Stavrakakis} Dr. John Stavrakakis has strong interests in 3D computer graphics, remote rendering and computer security. He holds a PhD in Computer science and is an academic fellow at the University of Sydney, Australia. 
\end{IEEEbiography}

\begin{IEEEbiography}[{\includegraphics[width=1in,height=1.25in,clip,keepaspectratio]{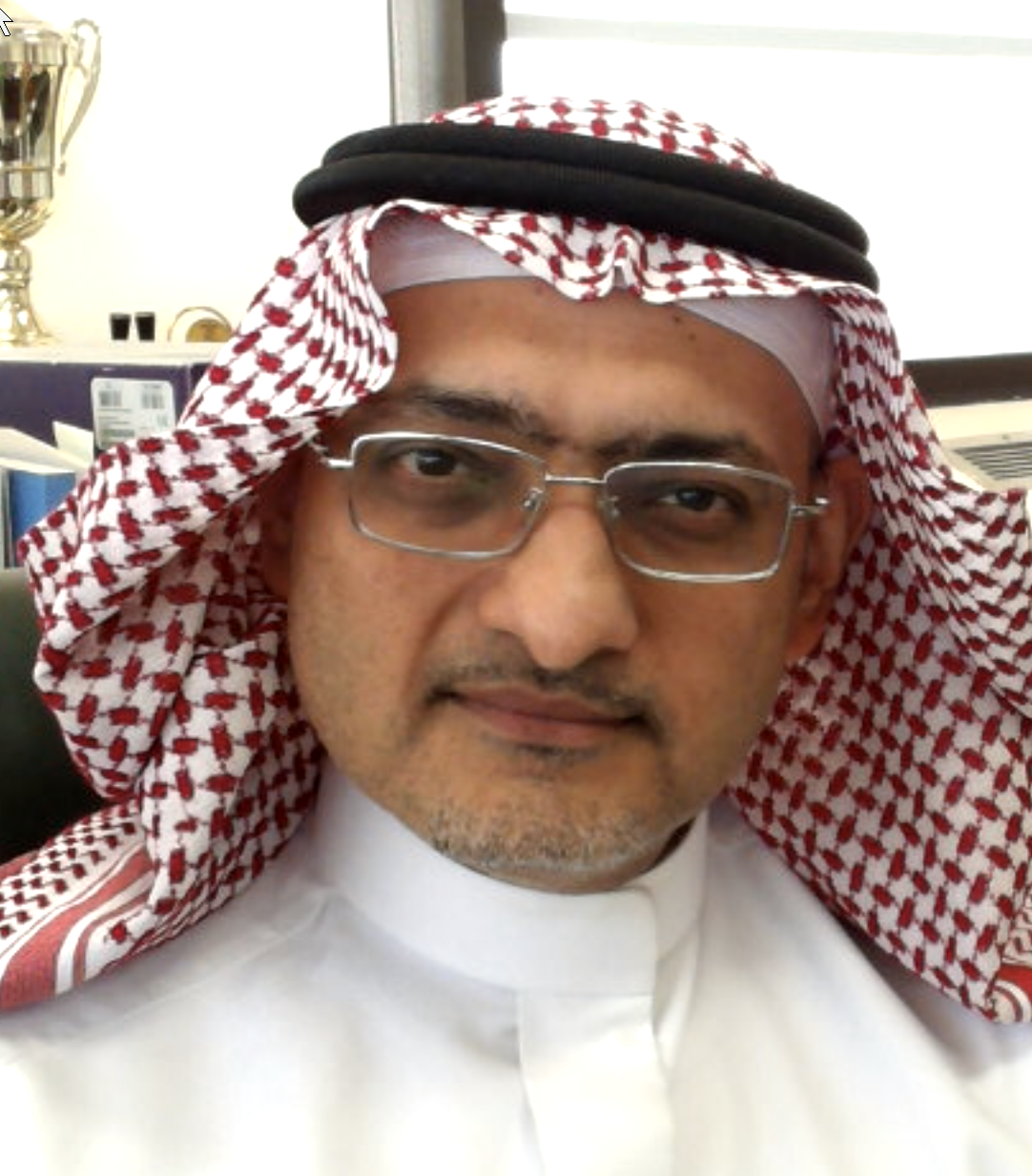}}]{Adel F. Ahmed} Dr. Adel F. Ahmed is an assistant professor at King Fahd University of Petroleum and Minerals. He received his PhD in computer science from the University of Sydney. His research interest includes Information Visualization and Machine Learning.
\end{IEEEbiography}

\begin{IEEEbiography}[{\includegraphics[width=1in,height=1.25in,clip,keepaspectratio]{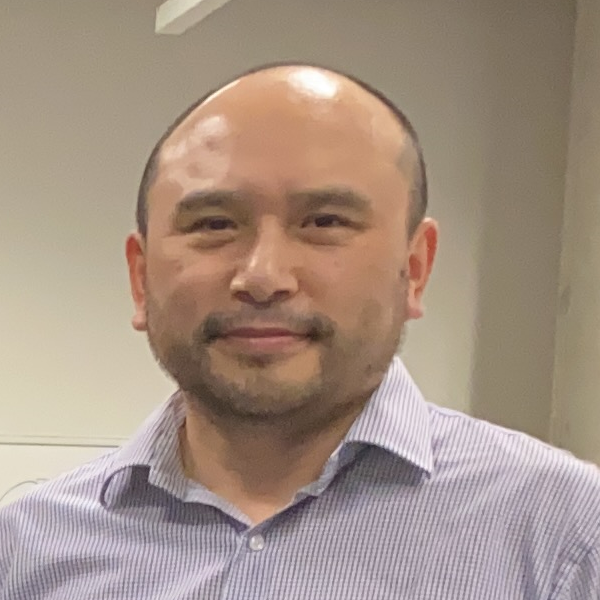}}]{Masahiro Takatsuka} Dr. Masahiro Takatsuka received his MEng degree at Tokyo Institute of Technology in 1992, and received his PhD at the Monash University in 1997. In 1997-2002, he worked at GeoVISTA Center, The Pennsylvania State University as a senior research associate. He joined the School of Computer Science, the University of Sydney in 2002.
\end{IEEEbiography}

\EOD

\end{document}